\documentclass{article}

     \usepackage[final]{neurips_2020}

\usepackage[utf8]{inputenc} 
\usepackage[T1]{fontenc}    
\usepackage{url}            
\usepackage{booktabs}       
\usepackage{amsfonts}       
\usepackage{nicefrac}       
\usepackage{microtype}      
\usepackage{geometry}
\usepackage{amsmath}
\usepackage{amssymb}
\usepackage{graphicx}
\usepackage{algorithm}
\usepackage{algpseudocode}
\usepackage{bm}
\usepackage[english]{babel}

\usepackage{wrapfig}

\usepackage{multirow}

\usepackage[dvipsnames]{xcolor}
\definecolor{darkblue}{rgb}{0, 0, 0.5}
\usepackage[colorlinks=true,linkcolor=darkblue,citecolor=darkblue,urlcolor=darkblue]{hyperref}

\newtheorem{thm}{Theorem}
\newtheorem{lem}{Lemma}
\newtheorem{asm}{Assumption}

\newcommand{\vv}[1]{\boldsymbol{#1}}
\global\long\def\th{\boldsymbol{\theta}}%
\global\long\def\bpm{\boldsymbol{\phi}_{\text{Matrix}}}%
\global\long\def\bp{\boldsymbol{\phi}}%

\global\long\def\x{\boldsymbol{x}}%
\global\long\def\z{\boldsymbol{z}}%

\global\long\def\DD{\mathbb{D}}%
\global\long\def\E{\mathbb{E}}%
\global\long\def\R{\mathbb{R}}%

\global\long\def\D{\mathcal{D}}%

\global\long\def\lip{\text{Lip}}%

\global\long\def\s{\mathtt{score}}%
\global\long\def\M{\mathcal{M}}%
\global\long\def\A{\boldsymbol{A}}%
\global\long\def\w{\boldsymbol{w}}%
\global\long\def\W{\boldsymbol{W}}%
\global\long\def\h{\boldsymbol{h}}%

\title{Greedy Optimization Provably Wins the Lottery: Logarithmic Number of Winning Tickets is Enough}

\author{%
  \large Mao Ye \thanks{Equal Contribution} \\
  UT Austin \\
  \texttt{my21@cs.utexas.edu} \\
  \And
  \large Lemeng Wu \textsuperscript{*} \\
  UT Austin\\
  \texttt{lmwu@cs.utexas.edu} \\
  \And
  \large Qiang Liu \\
  UT Austin \\
  \texttt{lqiang@cs.utexas.edu} \\
}

\newcommand{\norm}[1]{\left\lVert#1\right\rVert}

\DeclareMathOperator*{\argmin}{arg\,min}

\begin{document}

\maketitle

\begin{abstract}
Despite the great success of deep learning, recent works show that large deep neural networks are often highly redundant and can be significantly reduced in size. However, the theoretical question of how much we can prune a neural network given a specified tolerance of accuracy drop is still open. This paper provides one answer to this question by proposing a greedy optimization based pruning method. The proposed method has the guarantee that the discrepancy between the pruned network and the original network decays with exponentially fast rate w.r.t. the size of the pruned network, under weak assumptions that apply for most practical settings. Empirically, our method improves prior arts on pruning various network architectures including ResNet, MobilenetV2/V3 on ImageNet.

\end{abstract}

\section{Introduction}
Large-scale deep neural networks have achieved remarkable success on complex cognitive tasks, including image classification, e.g., \citet{he2016deep}, speech recognition, e.g., \citet{amodei2016deep} and machine translation, e.g.,\citet{wu2016google}. However, a drawback of the modern large-scale DNNs is their low inference speed and high energy cost, which makes it less appealing to deploy those models on edge devices such as mobile phones and Internet of Things \citep{cai2018proxylessnas}.

It has been shown that network pruning \citep{han2015learning} is an effective technique to reduce the size of the DNNs without a significant drop of accuracy. However, most existing works on network pruning are based on heuristics, leaving the theoretical questions largely open on what kind of network can be effectively pruned, how much we can prune a DNN   
given a specified tolerance of accuracy drop and how to achieve it with a practical and computationally efficient procedure.

Recently, a line of works on network pruning with theoretical guarantees have emerged,
including sensitivity-based methods \citep{baykal2019sipping, Liebenwein2020Provable}, coreset-based methods \citep{baykal2018datadependent,Mussay2020Data-Independent}, greedy forward selection \citep{ye2020good}. Both the sensitivity-based and coreset-based methods prune the network by sampling and bound the error caused pruning via concentration inequalities. They show that the error introduced by pruning decays with an $\mathcal{O}(n^{-1})$ rate w.r.t. the size $n$ of pruned network. 
This is comparable to the asymptotic error obtained by directly training a neural network of size $n$  with gradient descent descent, which is also $\mathcal{O}(n^{-1})$ following the mean field analysis of \citet{mei2018mean, araujo2019mean, sirignano2019mean}. More recently, \citet{ye2020good} proposed the first pruning method that achieves a faster $\mathcal{O}(n^{-2})$ error rate and is hence provably better than direct training with gradient descent. 
See Table \ref{tbl: rate_compare} for a summary on those works.

However, the analysis of 
\citet{ye2020good} only applies to two-layer  networks and requires the original network to be sufficiently over-parameterized. 
In this paper, we proposed a new greedy optimization based pruning method, which learns sub-networks of size $n$ with a significantly smaller $\mathcal{O}(\exp(- cn))$ error rate, 
 improving the rate from polynomial to exponential. 
 In addition, our theoretical rate only requires weak assumptions that hold for most networks in practice, without requiring the the original networks to be overparameterized as \citet{ye2020good}. Different from the Lottery Ticket Hypothesis \citep{frankle2018lottery}, which selects the winning tickets that give good performance when trained in isolation from initialization, our approach finds the tickets (that already won) from a fully converged network.

 Practically, our algorithm is simple and easy to implement. In addition, we introduce practical speedup techniques to further improve the time efficiency. 
 Empirically, our method improves the prior arts  on network pruning under various network structures including ResNet-34 \citep{he2016deep}, MobileNetV2 \citep{sandler2018mobilenetv2} and MobileNetV3 \citep{howard2019searching} on ImageNet \citep{deng2009imagenet} as well as DGCNN \citep{wang2019dynamic} on ModelNet40 \citep{wu20153d} on point cloud classification.

\begin{table} \label{tbl: rate_compare}
\begin{centering}
\begin{tabular} {c|ccc}
\hline 
 & Rate & No Over-param & Deep Net\tabularnewline
\hline 
\citet{baykal2019sipping, Liebenwein2020Provable} & $\mathcal{O}(n^{-1})$ & $\checkmark$ & $\checkmark$\tabularnewline
\citet{baykal2018datadependent,Mussay2020Data-Independent} & $\mathcal{O}(n^{-1})$ & $\checkmark$ & $\checkmark$\tabularnewline
\citet{ye2020good} & $\mathcal{O}(n^{-2})$ & $\times$ & $\times$\tabularnewline
This paper & $\mathcal{O}(\exp(-cn))$ & $\checkmark$ & $\checkmark$\tabularnewline
\hline 
\end{tabular}\caption{Overview on theoretical guaranteed pruning methods. Rate above gives how the error due to pruning decays as the size of the pruned network ($n$) increases. Column `No Over-param' denotes whether the method applies to an original network that is not over-parameterized in order to obtained the rate. Column `Deep net' denotes whether the analysis applies to deep networks.}
\par\end{centering}
\end{table}

\paragraph{Notation}
We use notation $[N]:={1,...,N}$ for the set of the first $N$ positive integers. All the vector norms $\left\Vert \cdot\right\Vert$ are assumed to be $\ell_2$ norm. We denote the vector $\ell_{0}$ norm by $\left\Vert \cdot\right\Vert _{0}$. $\left\Vert \cdot\right\Vert _{\lip}$ denotes the Lipschitz norm for functions. $\mathbb{I}\{\cdot\}$ indicates the indicator function.

\newcommand{\aiell}{{a_{\ell,i}}}

\section{Background and Method}
\paragraph{Problem Setup}
Given a pre-trained deep neural network with $L$ layers:
$F(\x)=F_{L}\circ F_{L-1}\circ\cdots \circ F_{2}\circ F_{1}(\x),$
where the $\ell$-th layer $F_\ell$ consisting of $N$ neurons forms a mapping of form \[
F_{\ell}(\z)=\frac{1}{N}\sum_{i=1}^{N}\sigma(\th_{i}^{\ell},\z),
\]
with $\z$ as a proper input of the $\ell$-th layer, which is the output of the previous $\ell-1$ layers. 
Here $\sigma(\th,\cdot)$ is a general nonlinear map parameterized by $\th$ 
that represents a neuron or other module in the network. 
For example, in a fully connected layer, we have $\sigma(\th,\z)=w_{1}\sigma_{+}(\boldsymbol{w}_{2}^{\top}\z)$ with  $\th=[w_{1},\boldsymbol{w}_{2}]$ and $\sigma_+$ an activation function such as ReLU or Tanh. In a convolution layer,  we have $\sigma(\th,\z)=w_{1}\sigma_{+}(\boldsymbol{w}_{2}*\z)$, where $*$ denotes the convolution operator. 
In this paper we may call $\sigma(\th_i^\ell,\cdot)$ the neuron $i$ or the $i$-th neuron for simplicity. 
Without loss of generality,
we assume each layer in the given deep network has the same number of neurons using the same activation function. 

The goal of network pruning is 
to construct a thinner network by replacing each layer with a subset of $n<N$ neurons. For simplicity of presentation, we focus on pruning a single layer $F_\ell$ for now and we discuss how to apply our algorithm in a layer-wise fashion to prune the whole network in section \ref{sec: prac}.

To prune the $\ell$-th layer, 
the goal is to replace $F_\ell$  
with a thinner layer $f_{\ell,A}$ with $n<N$ neurons: 
\[
f_{\ell,\A}(\z)=\sum_{i=1}^{N}a_{i}\sigma(\th_{i}^{\ell},\z), \ \ \A=[a_1,...,a_N]\in\Omega_{N},\left\Vert \A\right\Vert _{0}\le n,
\]
where $\Omega_{[N]}$ is the probability simplex on the $N$ neurons, that is, 
\[
\Omega_{N}=\Big\{\boldsymbol{v}:\boldsymbol{v}=[v_{1},...,v_{N}]\in\R^{N},~~~v_{i}\ge0,~~~\ \forall i\in[N]\ ~~~\text{and}\ ~~~ \sum_{i=1}^{N}v_{i}=1\Big\}.
\]
By enforcing that $\A \in\Omega_N$, we prune the layer by finding the best convex combination of a subset of neurons. The constraint that $\sum_{i}a_{i} = 1$ ensures that the overall magnitude of the output of the layer after pruning matches that of the original network even when a lot neurons are moved. 
We denote the network with the $\ell$-th layer replaced by $f_{\ell,A}$ as $f_A$, i.e., 
\[
f_{A}=F_{L}\circ...\circ F_{\ell+1}\circ f_{\ell,A}\circ F_{\ell-1}\circ...\circ F_{1}. 
\]

Given an observed dataset $\D_m:=(\x^{(i)}, y^{(i)})_{i=1}^m$ with $m$ data points.
We want to choose $\vv A$ such that the pruned network $f_{\vv A}$ is close to the original $F$ as much as possible, measured by the regression discrepancy loss,  
\[
\DD[f_{\vv A}, ~ F]=\E_{\x\sim\D_{m}}\left [\left(f_{A}(\x)-F(\x)\right)^{2} \right].
\]
Our algorithm and theoretical analysis can be extended to other discrepancy losses such as the cross-entropy. The problem of pruning the $\ell$-th layer can be formulated by the following constraint problem
\begin{eqnarray} \label{opt: all}
\min_{\A} \ \DD[f_{\A},F],\ \ \text{s.t.}
\ \   ~~~ \A \in \Omega_N, ~~~ 
\left\Vert \A \right\Vert_0 \le n.
\end{eqnarray}
This yields a challenging sparse optimization problem, which we address using   greedy optimization, yielding algorithms that are both theoretically guaranteed and practically efficient. 

\subsection{Pruning with Greedy Local Imitation} \label{sec: local imitation}
We first introduce a simply greedy algorithm via local imitation for searching a good solution of problem (\ref{opt: all}). The pruned network can be viewed as
\[
H\circ f_{\ell,\A}(\z) =H\circ\left(\sum_{i=1}^{N}a_{i}\sigma(\th_{i}^{\ell},\z)\right).
\]
Here $\z$ is the output of the $\ell-1$-th layers and $H=F_{L} \circ ...\circ F_{\ell+1}$ is the mapping of the later layers. Denote $\z^{(i)}=F_{\ell-1}\circ...\circ F_{1}(\x^{(i)})$. The set $\D_{m}^\ell:=(\z^{(i)})_{i=1}^m$ denotes the distribution of training data pushed through the first $\ell-1$ layers. Suppose $H$ is Lipschitz continuous, which typically holds for neural networks, we are about to upper bound $\DD[f_{\A},F]$ by $\DD[f_{\A},F]\le\left\Vert H\right\Vert _{\lip}^{2}\bar{\DD}[f_{\ell,\A},F_{\ell}]$, where $\bar{\DD}$ is the local discrepancy loss measuring the discrepancy on the output of the $\ell$-th layer between the pruned and original network
\[
\bar{\DD}[f_{\ell,\A},F_{\ell}]=\E_{\z\sim\D_{m}^\ell}\left\Vert f_{\ell,\A}(\z)-F_{\ell}(\z))\right\Vert ^{2}.
\]
In local imitation, we construct $f_{\ell,A}$ such that its output well imitates the output of $F_\ell$, i.e.,
\begin{eqnarray} \label{opt: local}
\min_{\A} \ \bar{\DD}[f_{\ell, \A},F_\ell],\ \ \text{s.t.}
\ \   ~~~ \A \in \Omega_N, ~~~ 
\left\Vert \A \right\Vert_0 \le n.
\end{eqnarray}
Importantly, different from the original loss $\DD[\cdot]$ in \eqref{opt: all}, the layer-wise local discrepancy loss $\bar{\DD}[\cdot]$  
is convex w.r.t. $\A$ and enjoys good geometric property for enabling fast exponential error rate via greedy optimization, as we show in sequel. On the other hand, as the final discrepancy $\DD$ is controlled by the local discrepancy $\bar{\DD}$, i.e., minimizing $\bar{\DD}$ effectively minimizes $\DD$.

The local imitation is a bi-directional greedy optimization for solving \eqref{opt: local}. It starts with an empty layer, and sequentially adds, removes or adjusts neurons that yield the largest decrease of the loss. 
Specifically,  
 denote by $f_{\ell, \A(k)}$ the layer we obtained at the $k$-th iteration with $\A(k)=[a_1(k),...,a_N(k)]$. We start with selecting the single best neuron that minimizes the loss: 
\begin{eqnarray} \label{equ: local_init}
f_{\ell,\vv A(0)} = \sigma(\vv \theta_{i_0^*}^\ell, \cdot), ~~~~~ 
\text{ with~~~ $i_{0}^{*}=\underset{i\in[N]}{\arg\min}  ~\bar{\DD}[\sigma(\th_{i}^{\ell},\cdot),~~~F_{\ell}(\cdot)]$ }
\end{eqnarray}
where $i_0^*$ is the index of the selected neuron; correspondingly, we have 
$a_i(0)=\mathbb{I}\{i=i_{0}^{*}\}$. 

At iteration $k$, we search for the best neuron $i^*_k$ and step size $\gamma^*_k$ that minimizes the loss most, i.e.,
\begin{eqnarray} \label{equ: local_search}
[i_k^*, \gamma_k^*]=\argmin_{i\in [N], \gamma\in U_i} \bar{\DD}\big [(1-\gamma)f_{\ell,\A(k)}+\gamma\sigma(\th_{i}^\ell,\cdot), ~~ F_{\ell}\big ].
\end{eqnarray} 
Then we update $f_{\ell, \A(k)}$ to $f_{\ell, \A(k+1)} = (1-\gamma_k^*)f_{\ell,\A(k)}+\gamma_k^*\sigma(\th_{i}^\ell,\cdot)$.
Here $U_i$ in (\ref{equ: local_search}) is the search interval of the step size $\gamma$. We set $U_{i}=[0,1]$ if the $i$-th neuron has not been selected yet (i.e., $a_i(k)=0$) and $U_{i}=[{-a_i(k)}/({1-a_i(k)}),1]$ if the neuron has already been added (i.e., $a_i(k)>0$). Therefore, this update 
can correspond to adding or removing a neuron, or simply adjusting the weight of existing neurons: 
the $i_k^*$-th neuron is added into the pruned network $f_{\ell, \A}$ if we have $a_{i_k^*}(k)=0$, and it is removed from $f_{\ell, \A}$ if we have  $\gamma_k^*= -a_{i_k^*}(k)/({1-a_{i_k^*}(k)})$; no new neuron is added or removed if otherwise.

We stop the iteration when a convergence criterion, i.e., $\bar{\DD}\le\epsilon$, is met.

\paragraph{Solving  Greedy Optimization in \eqref{equ: local_search}}
A naive way to solve problem \eqref{equ: local_search} is by 
enumerating each neuron and solving the corresponding inner minimization on $\gamma$. This is computational costly as it requires computing the forward pass in neural network many times. However, given $i$, the local discrepancy loss is a quadratic function w.r.t. $\gamma$. Combined with some special property of the local imitation algorithms, we are able to solve \eqref{equ: local_search} with only computing the forward pass in network once. We refer readers to Appendix \ref{apx: solve local} for details

\subsubsection{Greedy Local Imitation Decays Error Exponentially Fast}

Now we proceed to give the convergence rate for the proposed local imitation algorithm. We introduce the following assumption.

\begin{asm} \label{asm: bound}
Assume that for any $i\in[N]$, $\z^{(j)} \in \D_m^\ell$, 
we have 
$\left\Vert \sigma(\th_{i}^{\ell},\z^{(j)})\right\Vert \le c_{1}$
and $\left\Vert H\right\Vert _{\lip}\le c_{1}$ for some $c_{1}<\infty$.
\end{asm}
%
Assumption \ref{asm: bound} holds when network parameters and data are bounded and the activation is Lipschitz continuous, 
which is very mild and holds for most network in practice.  
The following theorem characterizes the convergence of local imitation showing that the error caused by pruning decays exponentially fast when the number of neurons in the pruned model increases.

\begin{thm} [\textbf{Convergence Rate}] \label{thm: converge_deep} Under assumption \ref{asm: bound}, 
at each step $k$ of the greedy optimization in \eqref{opt: local}, 
we obtain a layer with no more than $k$ neurons (i.e., $\norm{\A(k)}_0\leq k $), 
whose loss satisfies $\DD[f_{\A(k)}, F] \le \left\Vert H\right\Vert _{\lip}^{2} \bar\DD[f_{\ell, \A(k)},F_{\ell}]=\mathcal{O}\left(\exp(-\lambda_{\ell}k)\right)$, 
where $\lambda_\ell>0$ is a strictly positive constant. That is, the loss decays exponentially with the number of neurons in $f_{\ell,\A(k)}$. 

\end{thm}

\paragraph{Remark}
Minimizing $\gamma$ over $U_i$ can be viewed as line searching the optimal step size for adjusting neuron $i$. 
We may also consider to 
choose a proper fixed step size, e.g., $\gamma=1/(k+1)$ instead of line searching,  in which case the optimization in \eqref{opt: local} is simplified into 
\begin{align} \label{equ:1kgamma}
\min_{i\in[N]}\bar{\DD}\left [\left ({kf_{\ell,\A(k)}+\sigma(\th_{i}^{\ell},\cdot)}\right)/(k+1),~~F_{\ell} \right],
\end{align}
which can be shown to give an $\mathcal{O}(k^{-2})$ 
error at the $k$-th step under the same assumption as Theorem~\ref{thm: converge_deep}. See Appendix \ref{apx: local fix step} for more details. 


\subsection{Pruning with Greedy Global Imitation} \label{sec: global imitation}
The local imitation method uses a surrogate local discrepancy loss which is convex w.r.t. $\A$ to prune the networks. Despite the good property of local imitation, the use of surrogate loss can be ineffective for some layers. 
For example, during the iteration of local imitation, the best neuron that minimizes the surrogate loss is not necessarily the best one that minimizes the actual discrepancy loss. 

We propose a second pruning method, which directly minimizing the original discrepancy loss. This method follows the similar greedy fashion as the local imitation. We initialize the network by 
$
f_{\A(0)} = H_2 \circ f_{\ell,\A(0)} \circ H_1 = H_2 \circ (\sum_{i=1}^{N}a_i(0)\sigma(\th_{i}^{\ell},\cdot)) \circ H_1,
$
where 
\begin{eqnarray} \label{equ: global_init}
a_i(0)=\mathbb{I}\{i=i_{0}^{*}\}, \ \ \ i_{0}^{*}=\underset{i\in[N]}{\arg\min}\ \DD[H_2\circ\sigma(\th_{i}^{\ell},\cdot)\circ H_1,F],
\end{eqnarray}
and $H_1=F_{\ell-1}\circ \cdots\circ F_1$ and $H_2 := F_L \circ \cdots \circ F_{\ell+1}$. 
Similarly, at each iteration, We adjust the network in a greedy way by solving the following problem:
\begin{eqnarray} \label{opt: global_ls}
\min_{i\in([N]}\min_{\gamma\in U_{i}}\DD\left[H_2\circ\left(\left(1-\gamma\right)f_{\ell,\A(k)}+\gamma\sigma(\th_{i}^{\ell},\cdot)\right)\circ H_1,F\right].
\end{eqnarray}
However, solving problem (\ref{opt: global_ls}) is computationally costly as the loss is non-convex w.r.t. $\A$ and thus solving the inner minimization on $\gamma$ requires exhaustive search. To reduce the computational cost, in iteration $k$, we instead consider the following problem 
\begin{eqnarray} \label{opt: global_herd}
\min_{i\in[N]}\DD\left[H_2 \circ\left(\left(1-\gamma_{k}\right)f_{\ell,\A(k)}+\gamma_{k}\sigma(\th_{i}^{\ell},\cdot)\right)\circ H_1,F\right],\ \ \gamma_{k}=(1+k)^{-1}.
\end{eqnarray}
Suppose that $i^*_k$ gives that solution of problem (\ref{opt: global_herd}), we update the network by setting 
\[
a_i(k+1)=(1-\gamma_{k})a_i(k)+\gamma_{k}\mathbb{I}\{i=i_{k}^{*}\}.
\]
And we end the iteration when convergence criterion is met. Notice that different from local imitation, the algorithm adjusts $A$ based on the final output of the network instead of the `local' output of the pruned layer and thus we name it global imitation.

Different from the local imitation, due to the nonlinear of $H_2$, besides Assumption \ref{asm: bound}, obtaining a convergence rate for the global imitation requires several additional assumptions characterizing the linearity of $H_2$ as well as the geometric property of the pruned layer. 
\begin{thm}
Under Assumption \ref{asm: bound} and some additional assumptions, specified in Appendix \ref{apx: global theory}, on the linearity of $H_2$ and initialization, we have $\DD[f_{\A(k)},F]=\mathcal{O}(k^{-2})$ and $\left\Vert \A(k)\right\Vert _{0}\le k$.
\end{thm}
\subsubsection{Accelerating Global Imitation via Taylor Approximation} \label{sec: taylor}
A native way to solve problem (\ref{opt: global_herd}) is by enumerating all the neurons and calculating $\DD\left[H\circ\left(\left(1-\gamma_{k}\right)f_{\ell,\A(k)}+\gamma_{k}\sigma(\th_{i}^{\ell},\cdot)\right),F\right]$, which has at least $\mathcal{O}(Nn)$ time complexity for pruning a layer with $N$ neurons to $n$ neurons. Here we propose a technique to reduce the computational cost via Taylor approximation. At iteration $k$, for any neuron $i\in[N]$, we have 
\[
\DD\left[H\circ\left[(1-\gamma_{k})f_{\ell,\A(k)}+\gamma_{k}\sigma(\th_{i}^{\ell},\cdot)\right],F\right]=\frac{1}{k+1}gr _{\A(k),i}+\mathcal{O}\left ((k+1)^{-2}\right),
\]
where we define
\[
gr_{\A(k),i}
=\frac{\partial}{\partial\gamma}\DD\left[H\circ\left[(1-\gamma)f_{\ell,A(k)}+\gamma\sigma(\th_{i},\cdot)\right],F\right] \bigg|_{\gamma=0}.
\]

Thus, when $k$ is large enough (which we find $25$ is sufficient
in practice), this approximation allows us to find the (near) optimal solution with small error of problem (\ref{opt: global_herd}) by finding the neuron with the largest $gr_{\A,i}$. Simple algebra shows that 
\begin{align*}
    gr_{\A,i}&=2\sum_{j=1}^{n}\left(\mathbb{I}\{j=i\}-a_{j}\right)r_{\A,i}, \ \ \ \text{where~~~}
    \\
    r_{\A,i}&:=\E_{\z\sim\D_{m}^{\ell}}\left[\left(H\circ f_{\ell,\A}(\z)-H\circ F_{\ell}(\z)\right)H'(f_{\ell,\A}(\z))\sigma(\th_{i}^{\ell},\z)\right].
\end{align*}
Therefore, we can easily calculate  $gr_{\A(k),i}$ for all $i\in [N]$  
once we obtain $r_{\A(k),i}$ for all $i\in[N]$. 
In appendix, we show that $r_{\A(k),i}$ can be easily computed with automatic differentiation function in common deep learning libraries by introducing some ancillary parameters into the model. See Appendix \ref{apx: taylor} for details. If we choose to use this approximation when $k>\tilde{k}$ for some $\tilde{k}>0$, we reduce the complexity from $\mathcal{O}(Nn)$ to $\mathcal{O}(n)$.

\subsection{Pruning vs GD: Numerical Verification of the Rate}
\begin{wrapfigure}{r}{.35\textwidth} 
\begin{centering}
\includegraphics[scale=0.35]{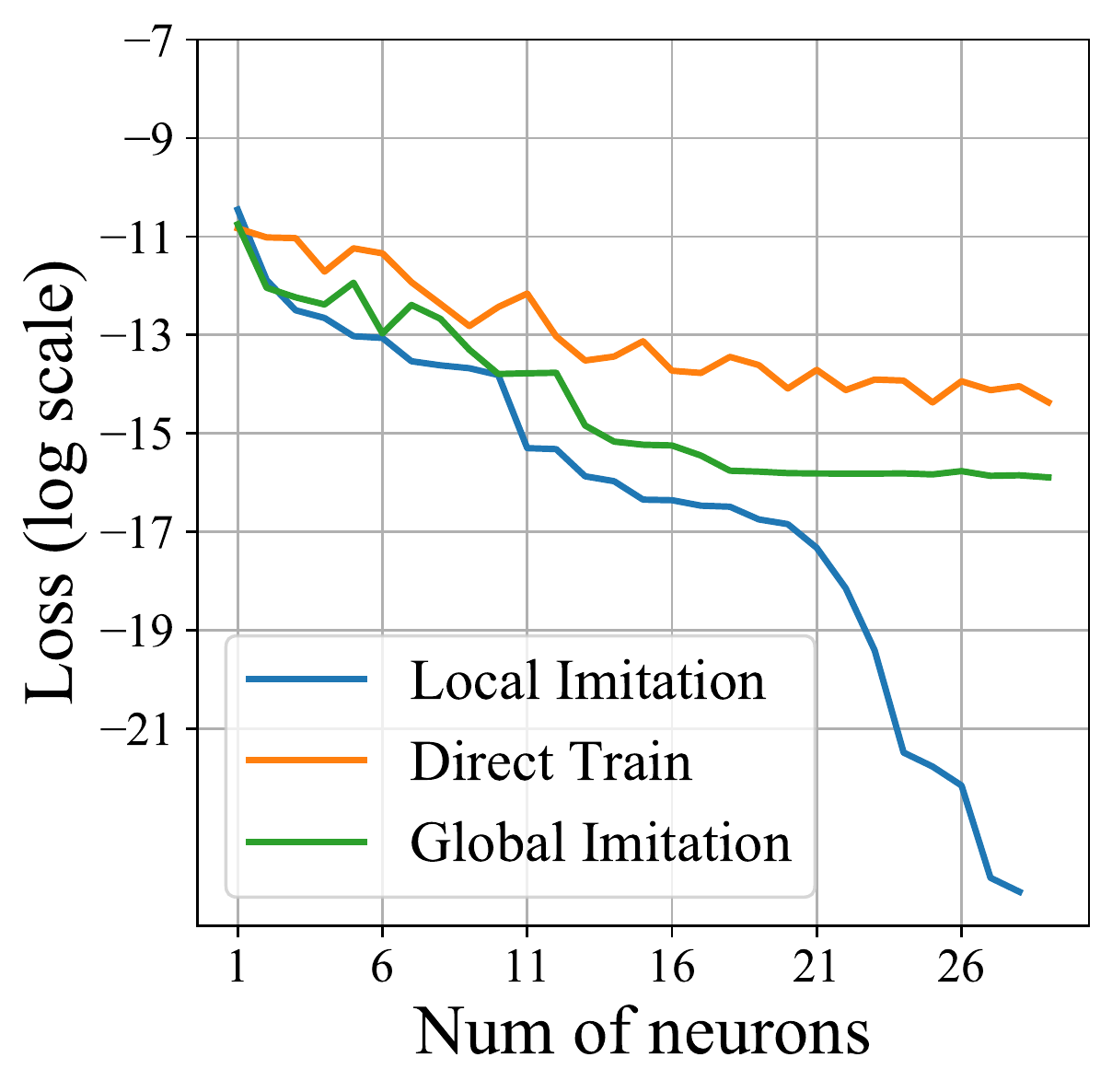}
\par\end{centering}
\caption{Discrepancy loss of the pruned model and train-from-scratch network with different sizes. The loss is in logarithm scale.} 
\label{fig: toy}
\end{wrapfigure}

Our result implies that the subnetwork $f_{\A}$ obtained by pruning gives $\DD[f_{\A},F]=\mathcal{O}\left(\exp(-\lambda n)\right)$ where $n$ is the number of neurons remained in the pruned layer. In comparison, the mean field analysis \citep{araujo2019mean, mei2018mean} suggests that directly train a network with same size as the pruned model gives $\mathcal{O}\left(n^{-1}\right)$ discrepancy loss. This suggests that pruning is provably better than training. We conduct a numerical experiment to verify the theoretical result. Given some simulated dataset, we firstly train a two hidden layer neural network with 100 neurons for each layer. And we prune the layer close to the input to different number of neurons using the local and global imitation. We also train the network with different number of neurons for the pruned layer and 100 neurons for the other one. Figure \ref{fig: toy} plots the discrepancy loss and the number of neurons of the pruned layer. The empirical result matches our theoretical findings. We refer readers to Appendix \ref{apx: rate verify} for more details.
\\
\\

\subsection{Practical Algorithm: Pruning All Layers} \label{sec: prac}
In section \ref{sec: local imitation} and \ref{sec: global imitation} we introduce how to use the greedy local/global imitation to prune a certain layer in a network. In order to prune the whole network, we apply the greedy optimization scheme in a layer-wise fashion. Starting from the full network $F$, we apply the pruning method to prune the first layer (the one that is closest to the input) $F_1$ to $f_1$, which returns a pruned network $F^{\text{prune},1}=F_{L}\circ F_{L-1}\circ\cdots\circ F_{2}\circ f_{1}$. We then apply the pruning method to prune the second layer $F_2$ in $F^{\text{prune},1}$ and continue until all the layers are pruned. By applying the pruning algorithm in this manner we prune the whole network.

The local imitation and global imitation perform differently when pruning different layers. To combine their advantages, when pruning each layer, both methods are applied individually with same convergence criterion and the one gives better performance is picked up. In this paper, we stop pruning when the discrepancy loss $\DD$ of the pruned model is smaller than a user specified threshold. The method prunes more neurons at convergence is selected. If both methods prune the same number of neurons, then the one with smaller discrepancy loss is chosen. Algorithm \ref{alg: local/global} summarizes the procedure of local and global imitation and Algorithm \ref{alg: all} gives the layer-wise scheme on pruning the whole deep network.

The exponential decay rate can be obtained by iteratively applying our theory on each layer. Suppose the pruned model $F^{\text{pruned}}$ has $n$ neurons at each layer, we have $\DD[F^{\text{pruned}}, F]=\mathcal{O}(\exp(-\lambda n))$. We refer reader to Appendix \ref{apx: rate all layer} for details. Also notice that the exponential decay rate also holds for Algorithm \ref{alg: local/global} as it chooses the method with smaller loss.

\begin{algorithm}[t]
\begin{algorithmic}[1]
\State{\textbf{Input}: A pretrained network $F$ with $L$ layers. The targeted layer index $\ell$ for pruning. Method = $\in$ \{local, global\}}
\State{Initialize $f_{\ell,\A}$ using (\ref{equ: local_init}) for local imitation else (\ref{equ: global_init}) for global imitation.}
\While{convergence criterion is not met}
    \State{Randomly sample a mini-batch data $\hat{\D}$.}
    \State{Update $f_{\ell,\A}$ by solving (\ref{equ: local_search}) for local imitation or (\ref{opt: global_herd}) for global imitation, using data $\hat{\D}$.}
\EndWhile

\State{\textbf{Return:} The pruned layer $f_{\ell,\A}$.}
\end{algorithmic}\caption{The Greedy Local/Global Imitation} \label{alg: local/global}
\end{algorithm}

\begin{algorithm}[t]
\begin{algorithmic}[1]
\State{\textbf{Input}: pretrained network $F$ with $L$ layers.}
\For{$\ell = 1,2,...,L$}
    \State{Obtain the pruned layer $f_{\ell, \A^{\text{local}}}$ by local imitation on $F$ with target layer $\ell$.}
    \State{Obtain the pruned layer $f_{\ell, \A^{\text{global}}}$ by global imitation on $F$ with target layer $\ell$.}
    \State{Replace the $\ell$-th layer $F_\ell$ of $F$ with $f_{\ell, \A^{\text{local}}}$ if local imitation is better, else $f_{\ell, \A^{\text{global}}}$.}
\EndFor
\State{\textbf{Return:} The pruned network $F$.}
\end{algorithmic}\caption{Layer-wise Prune} \label{alg: all}
\end{algorithm}

\section{Experiment}

\subsection{Comparing the Local and Global Imitation}
Our first experiment aims to analyze the performance of local and global imitation for pruning deep neural network for image classification. We first apply both methods to a pretrained VGG-11 on CIFAR-10 dataset. We prune all the 8 convolution layers individual (when pruning one layer, the other layers remain unpruned) using both local and global imitation in order to compare these two methods side by side. Code for reproducing can be found at \url{https://github.com/lushleaf/Network-Pruning-Greedy-Forward-Selection}.

\paragraph{Settings}
The full network is trained with SGD optimizer with momentum 0.9. We use 128 batch size with initial learning rate 0.1 and train the model for 160 epochs. We decay the learning rate by 0.1 at the 80-th and the 120-th epochs. During pruning we use 128 batch size. We do not apply the Taylor approximation tricks to global imitation for this experiment. we use cross entropy between the pruned model and original model as discrepancy loss.

\begin{figure}[t]
\begin{centering}
\includegraphics[scale=0.3]{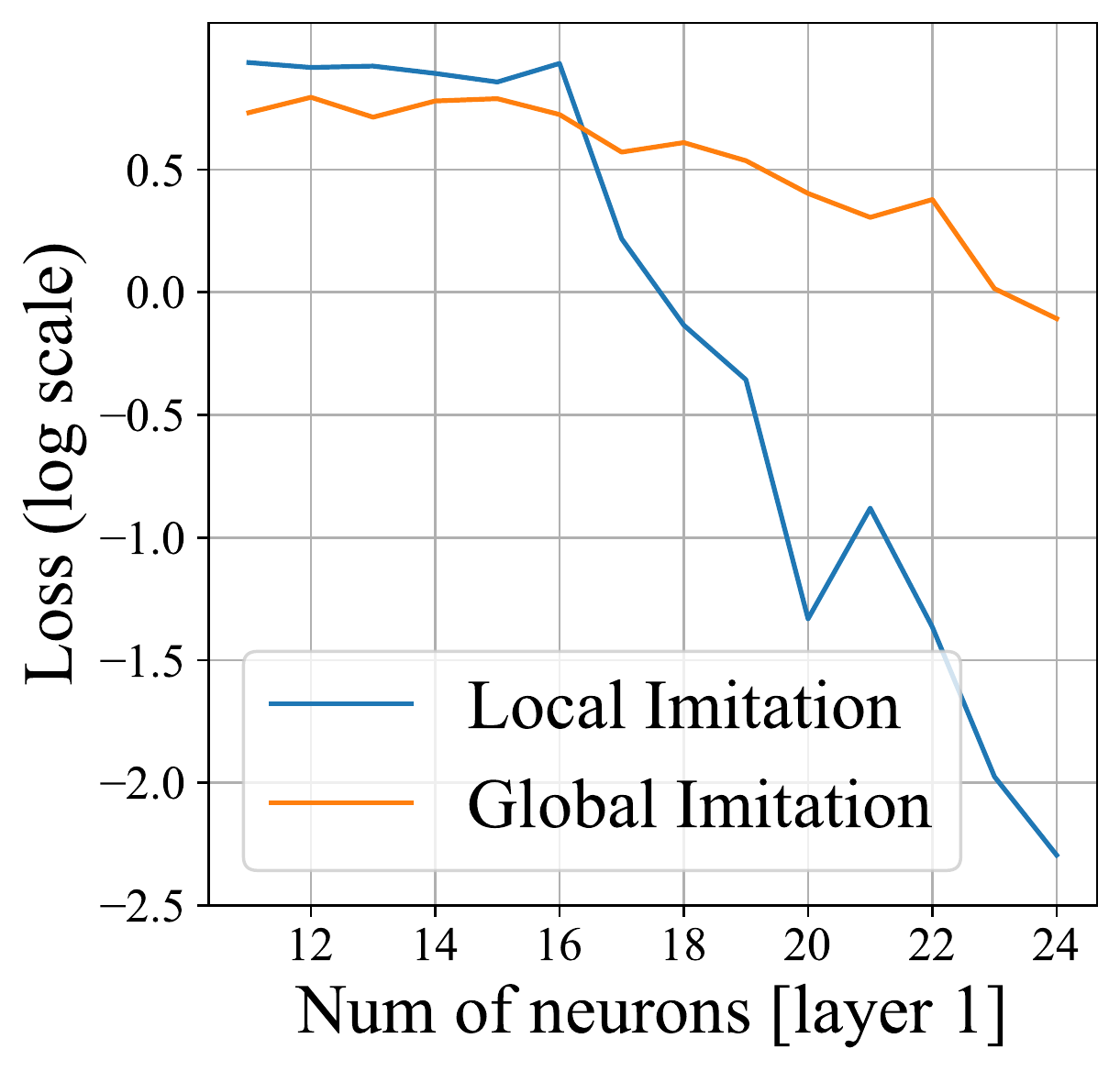}\includegraphics[scale=0.3]{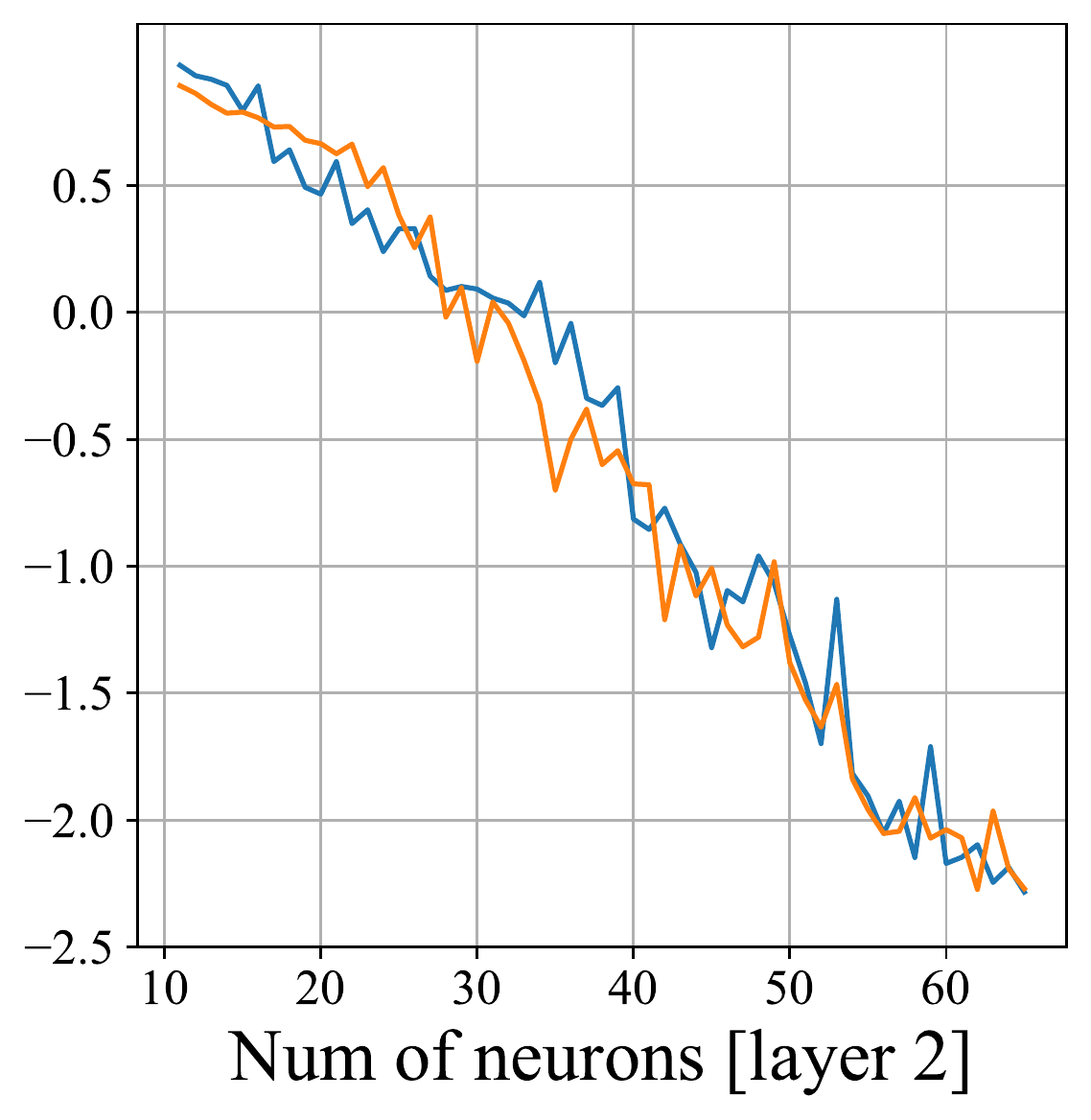}\includegraphics[scale=0.3]{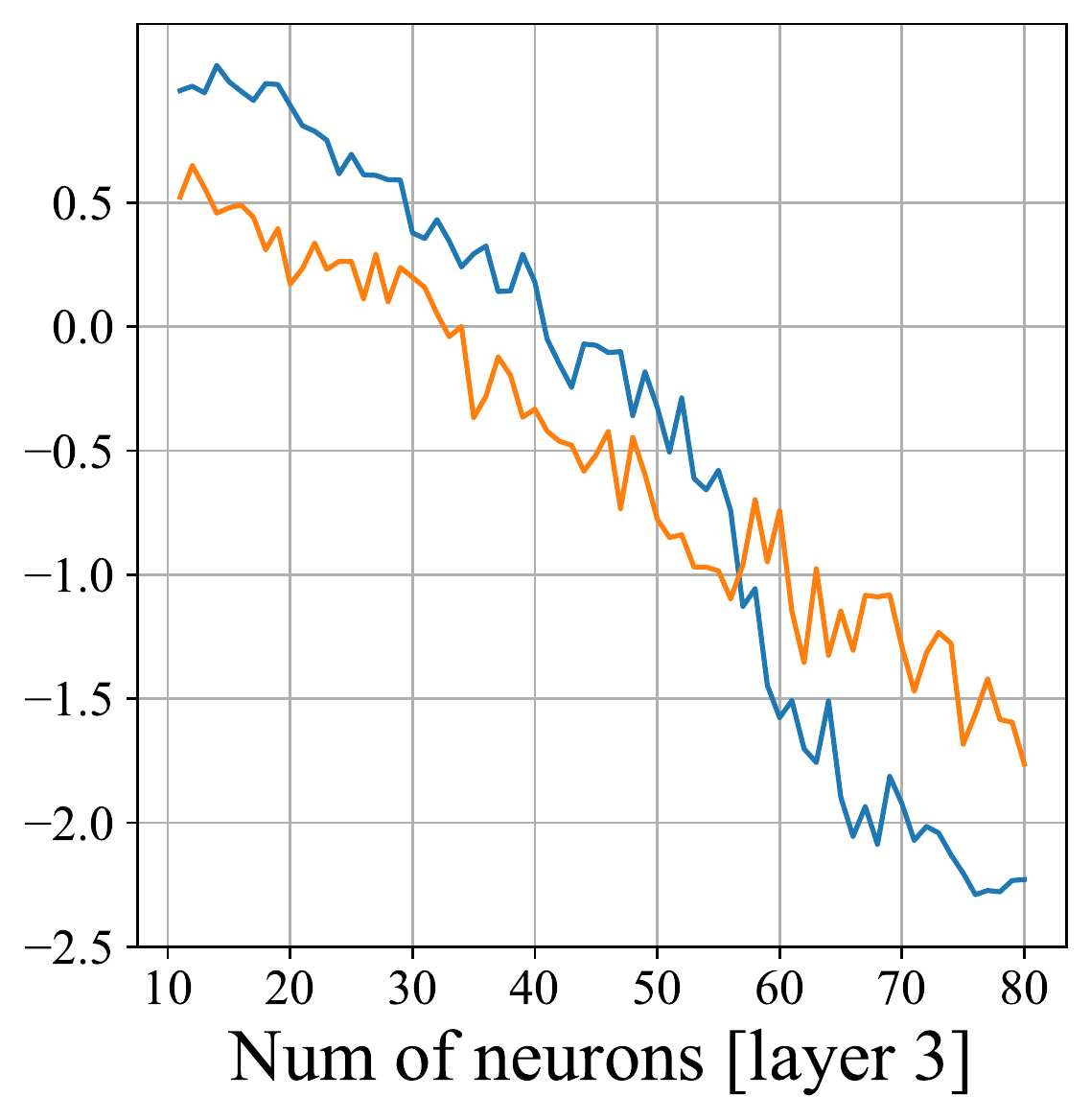}\includegraphics[scale=0.3]{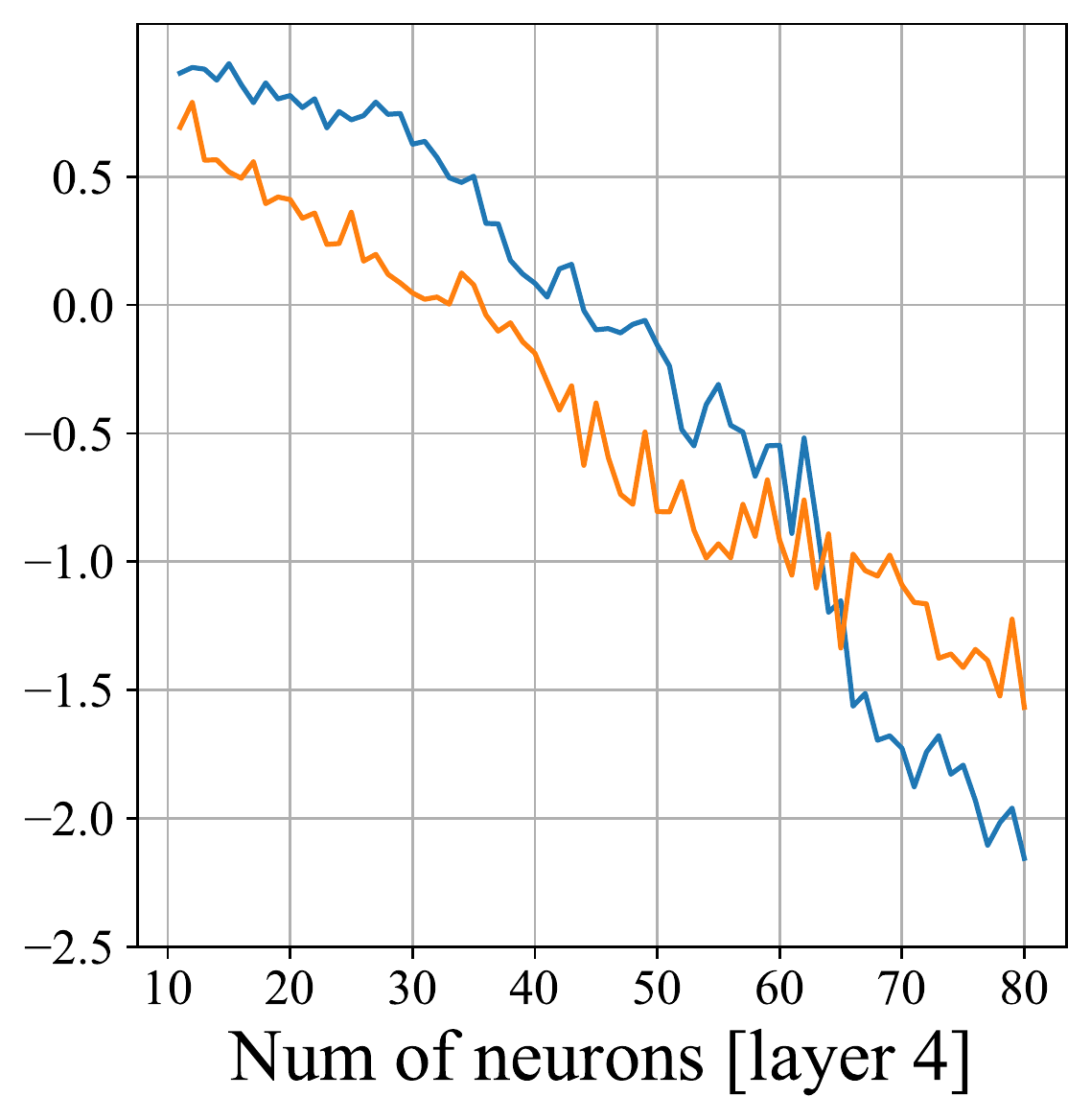}
\par\end{centering}
\begin{centering}
\includegraphics[scale=0.3]{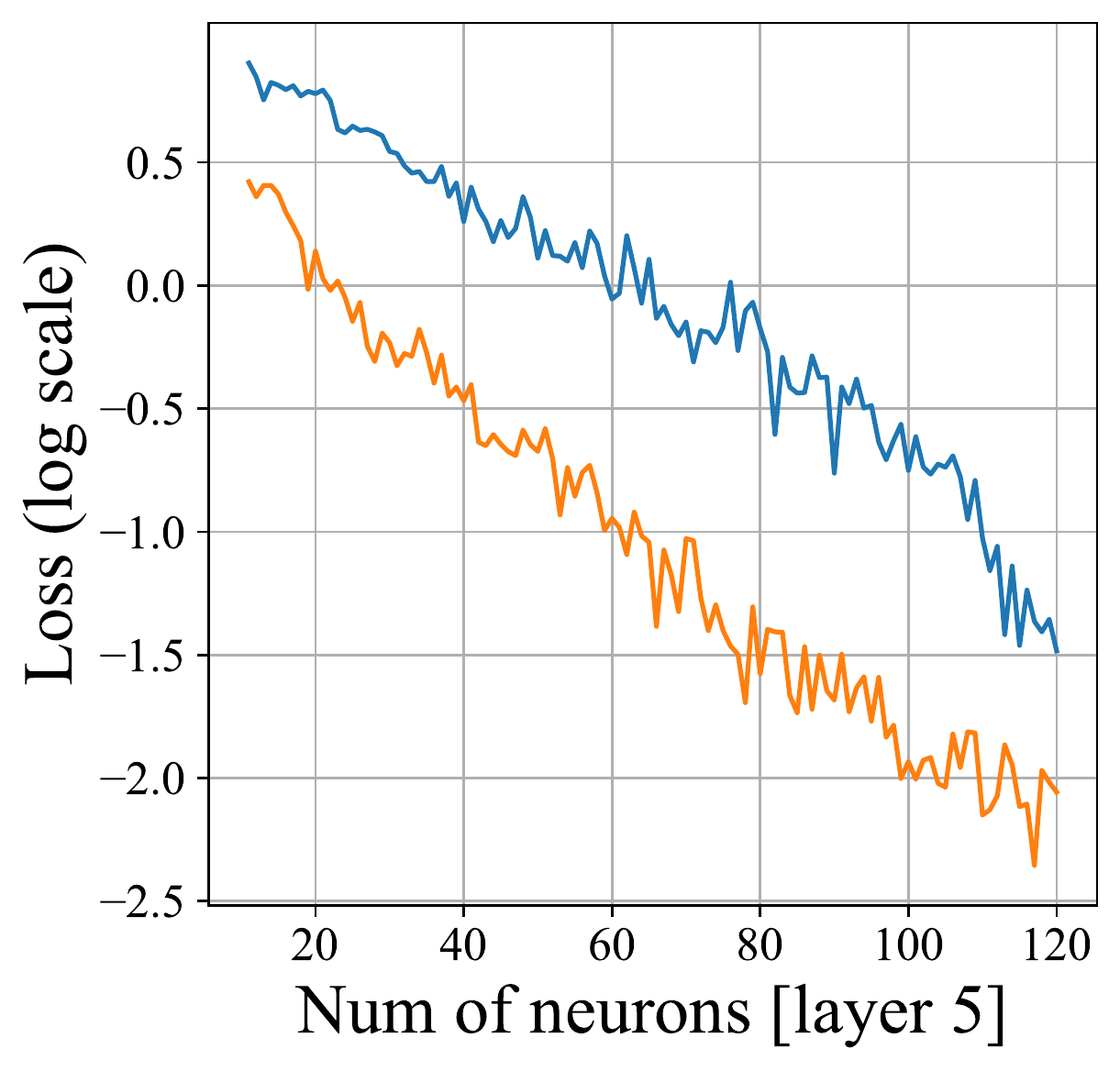}\includegraphics[scale=0.3]{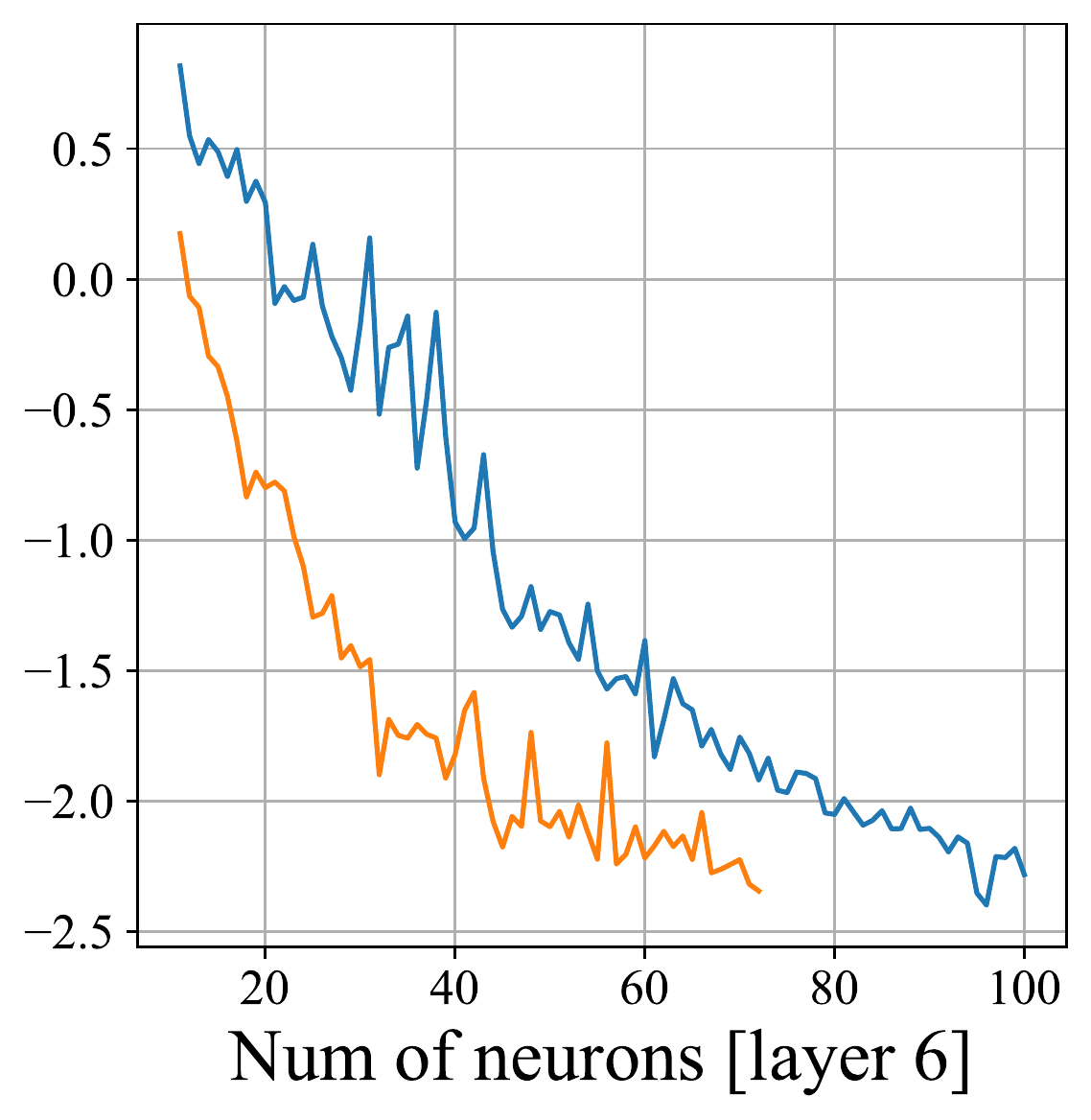}\includegraphics[scale=0.3]{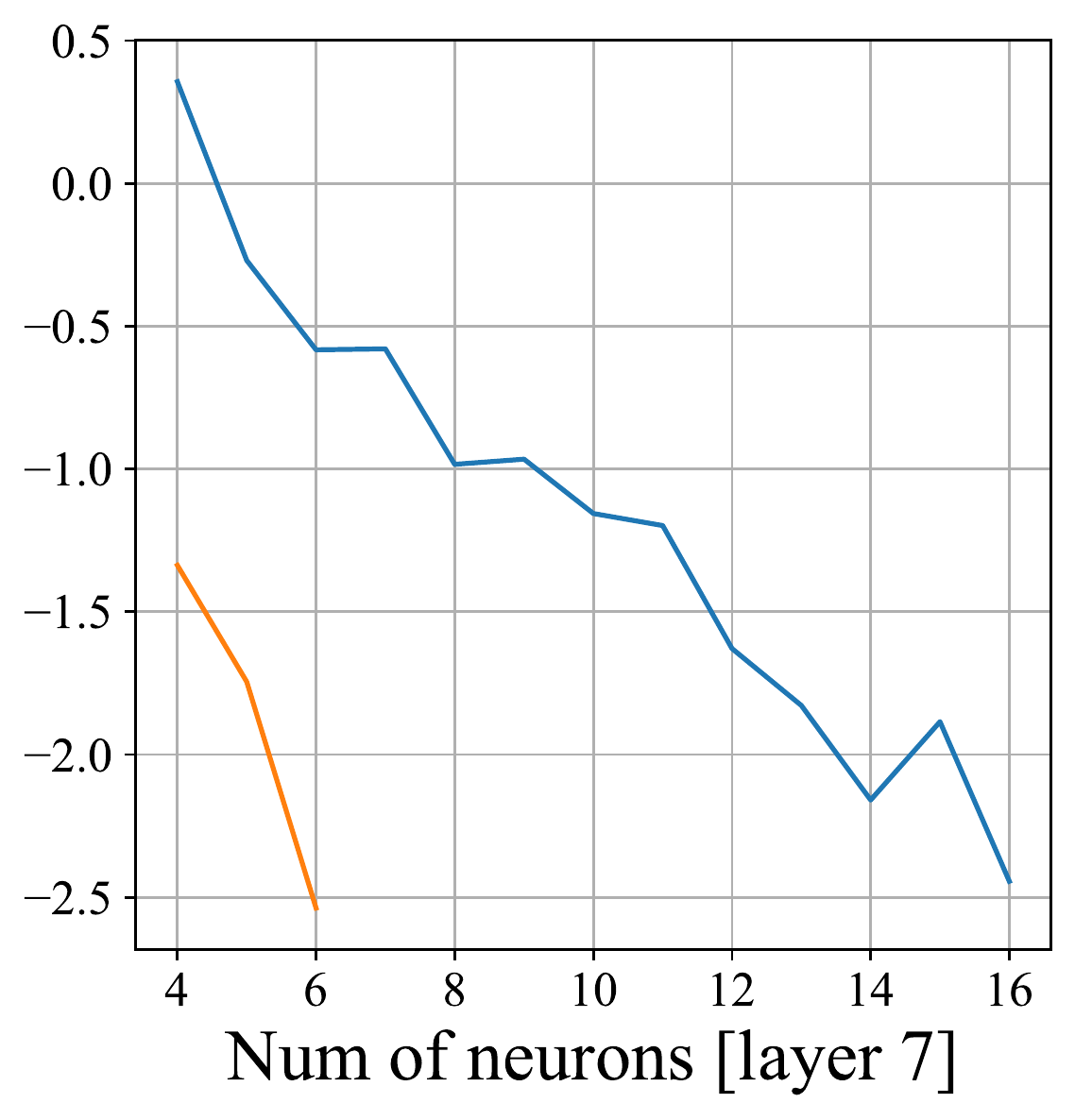}\includegraphics[scale=0.3]{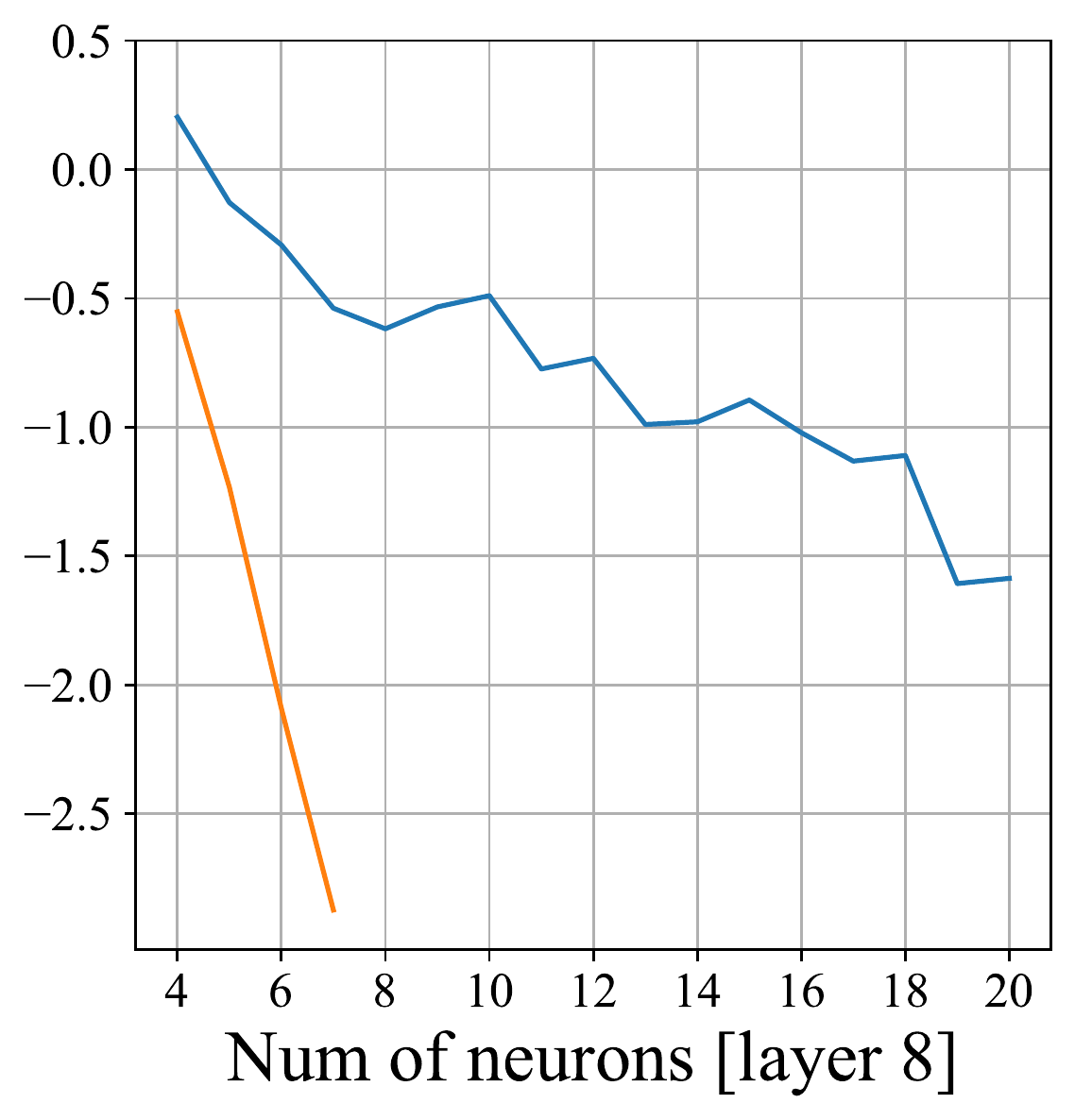}
\par\end{centering}
\caption{Pruning convolution layers on VGG11 using local and global imitation. From left to right and from top to bottom corresponds to the first (the one close to input) to the last convolution layers.} \label{fig: vgg11_local_global}
\end{figure}

\paragraph{Result}
Figure \ref{fig: vgg11_local_global} summarizes the result. Overall, we find the local and global imitation performs differently on different layers. The local imitation tends to decreases the loss faster on layer that is more close to input and with less neurons. While global imitation tends to performs much better than local imitation on layer that are close to output.

\paragraph{Combining Local and Global Imitation Outperforms Both}
In practice, we find purely pruning with local imitation tends gives worse result than pruning only with global imitation. However, combining the local imitation with global imitation performs better than pruning only with global imitation. To show this, we apply local and global imitation with the same setting as in section \ref{sec: prac} on pruning ResNet34 and MobileNetV2 on ImageNet. For comparison, pruning with only global imitation is also applied. The local+global setting achieves 73.5 top1 accuracy on the pruned ResNet34 with 2.2G FLOPs and 72.2 top1 accuracy on the pruned MobileNetV2 with 245M FLOPs. While pruning with only global imitation only achieve 73.2 top1 accuracy on ResNet34 and 72.1 top1 accuracy on MobileNetV2 with same FLOPs. The experimental settings are in Section \ref{sec: imagenet}.

\subsection{Imagenet Experiment} \label{sec: imagenet}
We use ILSVRC2012, a subset of ImageNet \citep{deng2009imagenet} which consists of about 1.28 million training images and 50000 validation images with 1000 different classes.

\textbf{Setting}
We apply our method on pruning ResNet34 (traditional large architecture) \citep{he2016deep}, MobileNetV2 (efficient architecture) \citep{sandler2018mobilenetv2} and MobileNetV3-small (an very small efficient architecture) \citep{howard2019searching} on ImageNet.

We use batch size 64 for both local and global imitation. We set the algorithm to converge when the gap between the cross entropy training loss before pruning and after pruning is smaller than $\epsilon$. We vary $\epsilon$ to get pruned model with different sizes. When conducting global imitation, we use the Taylor approximation trick introduced in section \ref{sec: taylor} to accelerate global imitation. We start the approximation when the number of neurons is larger than 25 and we evaluate the top 5 neurons with largest $gr_{\A,i}$ and pick up the best one to adjust. We find that this setting is able to produce the same pruning result as the exact version while substantially reduces the computation cost.

We finetune the pruned models with standard SGD optimizer with momentum 0.9 and weight decay $5\times10^{-5}$. All the pruned models are finetuned for 150 epochs with batch size 256 using cosine learning rate decay \citep{loshchilov2016sgdr}. We use initial learning 0.001 for ResNet34 and 0.01 for MobileNetV2 and MobileNetv3. We resize images to $224\times224$ resolution and adopt the standard data augumentation scheme (mirroring and shifting).

\textbf{Result} Table \ref{tbl: imagenet} reports the top1 accuracy, FLOPs and model size of the pruned network. Our algorithm consistently improves prior arts on network pruning.

\begin{table}[t]
\begin{centering}
\scalebox{1.0}{
\begin{tabular}{c|c|ccc}
\toprule[0.75pt]
Model & Method & Top-1 Acc & Size (M) & FLOPs\tabularnewline
\hline 
\multirow{10}{*}{ResNet34} & Full Model \citep{he2016deep} & 73.4 & 21.8 & 3.68G\tabularnewline
\cline{2-5} \cline{3-5} \cline{4-5} \cline{5-5} 
 & $L_1$ norm \citep{li2016pruning} & 72.1 & - & 2.79G\tabularnewline
 & Neural Imp \citep{molchanov2019importance} & 72.8  & - & 2.83G \tabularnewline
 & Rethink \citep{liu2018rethinking} & 72.9 & - & 2.79G\tabularnewline
 & More is Less \citep{dong2017more} & 73.0 & - & 2.75G\tabularnewline
 & GFS \citep{ye2020good} & 73.5 & 17.2 & 2.64G\tabularnewline
 & Ours & \textbf{73.5} & \textbf{14.9} & \textbf{2.20G}\tabularnewline
\cline{2-5} \cline{3-5} \cline{4-5} \cline{5-5} 
 & SPF \citep{he2018soft} & 71.8 & - & 2.17G\tabularnewline
 & FPGM \citep{he2019filter} & 72.5 & - & 2.16G\tabularnewline
 & GFS \citep{ye2020good} & 72.9 & 14.7 & 2.07G\tabularnewline
 & Ours & \textbf{73.3} & \textbf{13.5} & \textbf{1.90G}\tabularnewline
\hline 
\multirow{13}{*}{MobileNetV2} & Full Model \citep{sandler2018mobilenetv2} & 72.2 & 3.5 & 314M\tabularnewline
\cline{2-5} \cline{3-5} \cline{4-5} \cline{5-5} 
 & GFS \citep{ye2020good} & 71.9 & 3.2 & 258M\tabularnewline
 & Ours & \textbf{72.2} & \textbf{3.2} & \textbf{245M}\tabularnewline
 \cline{2-5} \cline{3-5} \cline{4-5} \cline{5-5} 
 & Uniform \citep{sandler2018mobilenetv2} & 70.4 & 2.9 & 220M\tabularnewline
 & AMC \citep{he2018amc} & 70.8 & 2.9 & 220M\tabularnewline
 & Meta Pruning \citep{liu2019metapruning} & 71.2 & - & 217M\tabularnewline
 & LeGR \citep{chin2019legr} & 71.4 & - & 224M\tabularnewline
 & GFS \citep{ye2020good} & 71.6 & 2.9 & 220M\tabularnewline
 & Ours & \textbf{71.7} & \textbf{2.9} & \textbf{218M}\tabularnewline
 \cline{2-5} \cline{3-5} \cline{4-5} \cline{5-5} 
& ThiNet \citep{luo2017thinet} & 68.6 & - & 175M\tabularnewline
& DPL \citep{zhuang2018discrimination} & 68.9& - & 175M\tabularnewline
 & GFS \citep{ye2020good} & 70.4 & 2.3 & 170M\tabularnewline
 & Ours & \textbf{70.5} & \textbf{2.3} & \textbf{170M}\tabularnewline
\hline 
\multirow{4}{*}{MobileNetV3-Small} & Full Model \citep{howard2019searching} & 67.5 & 2.5 & 64M\tabularnewline
\cline{2-5} \cline{3-5} \cline{4-5} \cline{5-5} 
 & Uniform \citep{howard2019searching} & 65.4 & 2.0 & 47M\tabularnewline
 & GFS \citep{ye2020good} & 65.8 & 2.0 & 49M\tabularnewline
 & Ours & \textbf{66.4} & \textbf{2.0} & \textbf{48M}\tabularnewline
\toprule[0.75pt]
\end{tabular}}\caption{Result on pruning deep neural networks on ImageNet.} \label{tbl: imagenet}
\par\end{centering}
\end{table}

\subsection{DGCNN Experiment}
We conduct experiment on the point cloud classification tasks on ModelNet40. Since the network structure used to extract the global information in point cloud usually requires to aggregate features from neighbor points, the high feature dimension heavily influence the forward time. We deploy our method on DGCNN. We compare with several baselines, including PointNet \citep{Qi_2017_CVPR}, PointNet++ \citep{NIPS2017pointnet++}, DGCNN with different width multipliers, and signed splitting steepest descent\citep{wu2020steepest}, which obtains a compact DGCNN by growing a extremely thin model. Table \ref{tab:dgcnn} shows that our method produces networks with comparable accuracy while with much less inference time. We refer readers to Appendix \ref{apx: dgcnn} for details of experiment settings.

\begin{table}[t]
\begin{center}

\scalebox{1.0}{
\begin{tabular}{l|ccc}
\hline
Model & Acc. & Forward time (ms) & \# Param (M)\\
\hline
PointNet \citep{Qi_2017_CVPR} & 89.2 & 32.19 & 2.85\\
PointNet++ \citep{NIPS2017pointnet++}& 90.7& 331.4 & 0.86\\
\hline
DGCNN (1.0x)& 92.6& 60.12 & 1.81\\
\hline
DGCNN (0.75x)& 92.4& 48.06 & 1.64\\
S3D \citep{wu2020steepest}& \textbf{92.9} & 42.06 & 1.51 \\
Ours & \textbf{92.9} & \textbf{37.43} & 1.49\\

\hline
DGCNN (0.5x)& 92.3& 38.90 & 1.52\\
Ours & \textbf{92.7} & \textbf{28.06} & 1.31\\
\hline
DGCNN (0.25x)& 91.8& 30.90 & 1.42\\
Ours & \textbf{92.5} & \textbf{24.06} & 1.24\\

\hline

\end{tabular}
}
\end{center}
\caption{Results on the ModelNet40 classification task.} 
\label{tab:dgcnn}
\vspace{-0.5em}
\end{table}

\section{Related Work}

\paragraph{Greedy Method}
Our method is highly related to \citet{ye2020good}, which is also a greedy method with $\mathcal{O}(n^{-2})$ error rate for pruning over-parameterized two layer network. In comparison, we obtain exponential decay rate for pruning deep neural network with no requirement on the over-parameterization of full model. Our local imitation method is also related to Frank Wolfe algorithm \citep{frank1956algorithm}. Compared with it, our local imitation is a bi-level greedy joint optimization method while Frank Wolfe first searches for 
best direction and then conduct descent greedily.

\paragraph{Theory on Lottery Ticket Hypothesis}
\citep{malach2020proving} aims to prove the existence of subnetwork inside a random network that well approximates an unknown target network which has finite width and depth. It shows that a sufficiently large random network (with a specific structure) contains such a subnetwork with width of higher order compared with the targeted network. Later \citet{pensia2020optimal,orseau2020logarithmic} improve the result by reducing the size of the original random network. \citet{elesedy2020lottery} also gives analysis of Lottery Ticket Hypothesis in linear model using the tool from compressive sensing. Compared with our method, their theoretical results require strong structure assumptions on the full model and pruned model. Besides, they fail to give an efficient algorithm to search for the subnetwork for deep learning model in practice. Notice that \citet{ye2020good} also gives analysis on Lottery Ticket Hypothesis and it is straightforward to combine their framework and our analysis to give faster rate.

\paragraph{Structured Pruning}
Existing methods on structured pruning includes the sparsity regularization based training methods, e.g., \citet{molchanov2017variational,liu2017learning,ye2018rethinking,huang2018data}; criterion based methods, e.g., \citet{molchanov2016pruning,li2016pruning,molchanov2019importance}, reconstruction error based method, e.g., \citet{he2017channel,luo2017thinet,zhuang2018discrimination,yu2018nisp} and direct search method, e.g., \citet{he2018amc,liu2019metapruning}. Our local imitation falls into the class of reconstruction error based method. Compared with those existing works, the proposed greedy optimization method enjoys good convergence property under weak assumptions and achieve better practical performance. \citet{zhou2020go} proposes a layer-wise imitation based training method for training deep and thin network, which is able to reduce the optimization error caused by the depth of the network. Their work is orthogonal to our work as we focus on reducing the error caused by small width.

\section{Conclusion}
This paper proposes a greedy optimization based pruning method, which is guaranteed to find a set of winning tickets (neurons) that approximates the fully trained unpruned network with exponential decay error rate w.r.t the number of selected tickets. The proposed pruning method is efficient with small time and space complexity and can be generally applied to various modern deep learning models.

\clearpage
\paragraph{Broader Impact Statement}
This work proposes a greedy optimization based pruning method, which has strong theoretical guarantee and good empirical performance. It gives positive improvement to the community of network efficiency. Our work do not have any negative societal impacts that we can foresee in the future.

\paragraph{Acknowledgement}
This paper is supported in part by NSF CAREER 1846421, SenSE 2037267 and EAGER 2041327.

\bibliographystyle{nips}
\bibliography{main}
\clearpage

\section*{Appendix}

We introduce the following extra notations which are used in several parts of the Appendix.
Suppose that $\th_{i}^{\ell},i\in[N]$ is the weight of the $N$ neurons
of the $\ell$-th layer in the original network $F$. We simplify
the notation by denoting $\th_{i}^{\ell}=\th_{i}$, $i\in[N]$. Suppose $\sigma(\th,\z)\in\R^{d}$, we define the data-dependent feature
by 
\[
\bpm\left(\th\right)=\left[\sigma(\th,\z^{(1)}),...,\sigma(\th,\z^{(m)})\right]^{\top}\in\R^{m\times d},
\]
and its vectorization 
\[
\bp(\th)=\left[\text{vec}^{\top}\left(\sigma(\th,\z^{(1)})\right),...,\text{vec}^{\top}\left(\sigma(\th,\z^{(m)})\right)\right]^{\top}\in\R^{md}.
\]
We also define $\h_{i}=\bp(\th_{i})$, $i\in[N]$, $\bar{\h}=\frac{1}{N}\sum_{i=1}^{N}\h_{i}$
and $\h_{\A}=\sum_{i=1}^{N}a_{i}h_{i}$ with $\A=[a_{1},a_{2},...,a_{N}]$.
Define $\M=\text{conv}\left(\left\{ \bp(\th_{i})\mid i\in[N]\right\} \right)$
as the convex hull generated by the set $\left\{ \bp(\th_{i})\mid i\in[N]\right\} $.
Given some set $M\subseteq\R^{d}$, we denote the relative interior
of $M$ by $\text{ri}M$, the closure of $M$ by $\text{cl}M$ and
the affine hull of $M$ by $\text{Aff}M$. We define $\mathcal{B}(\x_{0},r)$ as the ball centered at $\x_{0}$ with radius $r$.

\subsection{Details on Solving (\ref{equ: local_search}) for Local Imitation} \label{apx: solve local}

Now we describe the approach to solve problem \eqref{equ: local_search} with only compute one forward pass.
Define
\[
V_{i,\A}(\gamma)=\bar{\DD}[(1-\gamma)f_{\ell,\A}+\gamma\sigma(\th_{i}^{\ell},\cdot),F_{\ell}]=\gamma^{2}g_{i,\A}-2\gamma q_{i,\A}+\bar{\DD}[f_{\ell,\A},F_{\ell}],
\]
\begin{align*}
   \text{where ~} q_{i,\A}=\E_{\z\sim\D_{m}^{\ell}}\left[\left(F_{\ell}-f_{\ell,\A}\right)\left(\sigma(\th_{i}^{\ell},\cdot)-f_{\ell,\A}\right)\right],\ \ g_{i,\A}=\E_{\z\sim\D_{m}^{\ell}}\left[\left(\sigma(\th_{i}^{\ell},\cdot)-f_{\ell,\A}\right)^{2}\right].
\end{align*}
Notice that it is easy to obtain $\sigma(\th_{i}^{\ell},\z^{(j)})$ for all $i\in[N]$ and $j\in[m]$ by feeding the dataset into the neural network once, as it is the output of neuron $i$ in layer $j$. And thus $q_{i,\A(k)}$ and $g_{i,\A(k)}$ can be calculated cheaply given $\sigma(\th_{i}^{\ell},\z^{(j)}), i\in[N]$ and $j\in[m]$. Define $\tilde{\gamma}_{i,\A(k)}=q_{i,\A(k)}/g_{i,\A(k)}$, which is the optimum of $V_{i,\A(k)}(\gamma)$ w.r.t. $\gamma$ given $i$ (here the optimization of $\gamma$ is unconstrained). The following theorem shows some properties of the greedy local imitation method.

\begin{thm} \label{thm: solve_greedy_fast}
Under assumption \ref{asm: bound}, if $\bar{\DD}[f_{\ell,\A(k)}, F_\ell]>0$, then we have $\tilde{\gamma}_{\ell,i^*_{\ell,k}}<1$.
\end{thm}

Now we proceed to show how to obtain $i_{k}^{*}$ and $\gamma_{k}^{*}$
efficiently. Suppose at iteration $k$, $\bar{\DD}[f_{\ell,\A(k)},F_{\ell}]>0$
(otherwise the algorithm has converged). Given some neuron $i$ with
$a_{i}(k)=0$, we first calculate $\tilde{\gamma}_{i,\A(k)}$. If
$\tilde{\gamma}_{i,\A(k)}\in(0,1)$, then we define the score of this
neuron by the decrease of loss with this neuron selected, i.e.,
\[
\s_{\A(k)}(i):=\bar{\DD}[f_{\ell,\A},F_{\ell}]-\min_{\gamma\in U_{i}}V_{i,\A(k)}(\gamma)=-q^{2}_{i,\A(k)}/g_{i,\A(k)}.
\]
If $\tilde{\gamma}_{i,\A(k)}\ge1$, then from Theorem \ref{thm: solve_greedy_fast}, we
know that $i\neq i_{k}^{*}$. If $\tilde{\gamma}_{i,\A(k)}\le0$ and
if $i=i_{k}^{*}$, we have $\gamma_{k}^{*}=0$, which implies that
$\bar{\DD}[f_{\ell,\A(k+1)},F_{\ell}]=\bar{\DD}[f_{\ell,\A(k)},F_{\ell}]$.
It makes contradiction to Theorem \ref{thm: solve_greedy_fast}, which implies that $i\neq i_{k}^{*}$.
In this two cases, since neuron $i$ is not the optimal neuron to
select, we can safely set $\s_{\A(k)}(i)=0$. For neuron $i$ with
$a_{i}(k)>0$. Similarly, if $\tilde{\gamma}_{i,\A(k)}\ge1$, then
$i\neq i_{k}^{*}$ and thus we set $\s_{\A(k)}(i)=0$. If $\tilde{\gamma}_{i,\A(k)}\in U_{i}$,
then similarly $\s_{\A(k)}(i)=-q^{2}_{i,\A(k)}/g_{i,\A(k)}$. If $\tilde{\gamma}_{i,\A(k)}<-a_{i}(k)/(1-a_{i}(k))$,
then 
\[
\s_{\A(k)}(i)=V_{i,\A(k)}(-a_{i}(k)/(1-a_{i}(k))).
\]
And thus we have $i_{k}^{*}=\underset{i\in[N]}{\arg\max}\ \s_{\A(k)}(i)$.
Notice that the score of most neuron can be calculated cheaply using
$q_{i,\A(k)}$ and $g_{i,\A(k)}$. The only exception are neuron with
$a_{i}(k)>0$ and $\tilde{\gamma}_{i,\A(k)}<-a_{i}(k)/(1-a_{i}(k))$.
However, its score can be calculated using $\sigma(\th_{i}^{\ell},\cdot)$
and thus no extra forward pass is required.

\subsection{Local Imitation with Fixed Step Sizes} \label{apx: local fix step}
In this section we give detailed discussion on local imitation with a fixed step size scheme shown in Section 2.2.1. Different from the greedy optimization (\ref{equ: local_search}), in this scheme, as the step size is fixed, the solution returned in each iteration is no better than the one of (\ref{equ: local_search}). As a consequence, it gives slower convergence rate.

\begin{thm} \label{thm: local_fixstep_converge}
Under Assumption \ref{asm: bound}, at each step k of the greedy optimization in \ref{equ:1kgamma}, we have $\DD[f_{\A(k)},F]\le\left\Vert H\right\Vert _{\lip}^{2}\bar{\DD}\left[f_{\ell,\A(k)},F_{\ell}\right]=\mathcal{O}((k+1)^{-2})$ and $\left\Vert \A(k)\right\Vert _{0}\le k+1$.
\end{thm}

\subsection{Theory on Greedy Global Imitation} \label{apx: global theory}
Now we give the theoretical result on greedy global imitation. Denote $\kappa_{1}=\sup_{\A\in\Omega_{N}}\frac{\left\Vert H\circ\bar{\h}-H\circ\h_{\A}\right\Vert }{\left\Vert \bar{\h}-\circ\h_{\A}\right\Vert }$, $\kappa_{2}=\sup_{\A\in\Omega_{N}}\frac{\left\Vert \bar{\h}-\circ\h_{\A}\right\Vert }{\left\Vert H\circ\bar{\h}-H\circ\h_{\A}\right\Vert}$ and $D$ as the diameter of $\M$, which is defined in Lemma \ref{techlem: diameter}. Notice that $\kappa_1\kappa_2\ge1$. Using Lemma \ref{techlem: rl_avg}, we know that $\bar{\h}\in\text{ri}\M$,
which indicate that there exists some $\lambda>0$ such that 
\[
\mathcal{B}(\bar{\h},\lambda) \cap \text{Aff} \M \subseteq\M,
\]
where $\mathcal{B}(\bar{\h},\lambda)$ denotes the ball with radius
$\lambda$ centered at $\bar{\h}$.

\begin{thm} [Complete Version of Theorem 2] \label{thm: global_converge}
Suppose Assumption \ref{asm: bound} holds. Further suppose that 1. $D^{2}\ge\kappa_{1}^{2}\kappa_{2}^{2}(D^{2}-\lambda^{2})$; 2. at initialization $\left\Vert \bar{\h}-\h_{\A(0)}\right\Vert \le R$; 3. $\kappa_{1}D\le R$, where we define $R=\frac{\kappa_{1}^{2}\kappa_{2}\lambda+\kappa_{1}\sqrt{\kappa_{1}^{2}\kappa_{2}^{2}(\lambda^{2}-D^{2})+D^{2}}}{(\kappa_{1}^{2}\kappa_{2}^{2}-1)}$ ($R=+\infty$ if $\kappa_{1}\kappa_{2}=1$). Then we have $\DD[f_{\A(k)},F]=\mathcal{O}((k+1)^{-2})$, and $\left\Vert \A(k)\right\Vert _{0}\le k+1$.
\end{thm}

\paragraph{Remark}
Here the descending property of global imitation is influenced by the non-linear mapping $H$. As a consequence, the algorithm gives good convergence property when the whole dynamics is guaranteed to stay in a proper convergence region ($R$). The first extra assumption assumes the existence of this convergence region; The second extra assumption assumes a good initialization to ensure the dynamics stays in the convergence region at initialization; The third assumption can be roughly interpreted as assuming the dynamics will not jump out of the convergence region during descending. Notice that the extra assumptions holds when $\kappa_1$ and $\kappa_2$ is sufficiently close to 1.

\subsection{Details on Taylor Approximation Tricks} \label{apx: taylor}

In this section we give details on the computation of Taylor approximation
tricks. Notice that
\begin{align*}
gr_{\A,i} = & \frac{\partial}{\partial\gamma}\DD\left[H\circ\left[(1-\gamma)f_{\ell,\A}+\gamma\sigma(\th_{i},\cdot)\right],F\right]\bigg|_{\gamma=0}\\
 = & \frac{\partial}{\partial\gamma}\E_{\z\sim\D_{m}^{\ell}}\left(H\circ\left[(1-\gamma)f_{\ell,\A}(\z)+\gamma\sigma(\th_{i},\z)\right]-H\circ F_{\ell}(\z)\right)^{2}\bigg|_{\gamma=0}\\
 = & 2\E_{\z\sim\D_{m}^{\ell}}\left(H\circ\left[(1-\gamma)f_{\ell,\A}(\z)+\gamma\sigma(\th_{i},\z)\right]-H\circ F_{\ell}(\z)\right)\frac{\partial}{\partial\gamma}\left(H\circ\left[(1-\gamma)f_{\ell,\A}(\z)+\gamma\sigma(\th_{i},\z)\right]\right)\bigg|_{\gamma=0}\\
 = & 2\E_{\z\sim\D_{m}^{\ell}}\left(H\circ f_{\ell,\A}(\z)-H\circ F_{\ell}(\z)\right)H'\left(f_{\ell,\A}(\z)\right)\left(\sigma(\th_{i},\z)-f_{\ell,\A}(\z)\right)\\
 = & 2\E_{\z\sim\D_{m}^{\ell}}\left(H\circ f_{\ell,\A}(\z)-H\circ F_{\ell}(\z)\right)H'\left(f_{\ell,\A}(\z)\right)\left(\sigma(\th_{i},\z)-\sum_{j=1}^{N}a_{j}\sigma(\th_{j},\z)\right)\\
 = & 2\sum_{j=1}^{N}\left(\mathbb{I}\{j=i\}-a_{j}\right)r_{\A,j}.
\end{align*}
Thus the key quantities we want to obtain is 
\[
r_{\A,i}=\E_{\z\sim\D_{m}^{\ell}}\left[\left(H\circ f_{\ell,\A}(\z)-H\circ F_{\ell}(\z)\right)H'(f_{\ell,\A}(\z))\sigma(\th_{i}^{\ell},\z)\right].
\]
And once we obtain $r_{\A,i}$, we are able to calculate $gr_{\A,i}=2\sum_{j=1}^{N}\left(\mathbb{I}\left\{ j=i\right\} -a_{j}\right)r_{\A,j}$.
Now we introduce how to calculate $r_{\A,i}$ efficiently by introducing
an ancillary variable. Suppose when pruning layer $\ell$, we have
\[
f_{\ell,\A}(\z)=\sum_{i=1}^{N}a_{i}\sigma(\th_{i}^{\ell},\z)=\sum_{i=1}^{N}(a_{i}+b_{i})\sigma(\th_{i}^{\ell},\z),\ \ \text{where}\ b_{i}=0\ \forall i\in[N].
\]
Here $b_{i}$ is the introduced ancillary variable, which alway takes
$0$ value. We have 
\begin{align*}
 & \frac{\partial}{\partial b_{i}}\E_{\z\sim\D_{m}^{\ell}}\left(H\circ\left(\sum_{i=1}^{N}(a_{i}+b_{i})\sigma(\th_{i}^{\ell},\z)\right)-H\circ F_{\ell}(\z)\right)^{2}\bigg|_{b_{i}=0}\\
= & \E_{\z\sim\D_{m}^{\ell}}\left(H\circ f_{\ell,\A}(\z)-H\circ F_{\ell}(\z)\right)\frac{\partial}{\partial b_{i}}\left(H\circ f_{\ell,\A}\right)\bigg|_{b_{i}=0}\\
= & \E_{\z\sim\D_{m}^{\ell}}\left(H\circ f_{\ell,\A}(\z)-H\circ F_{\ell}(\z)\right)H'(f_{\ell,\A})\frac{\partial}{\partial b_{i}}f_{\ell,\A}\bigg|_{b_{i}=0}\\
= & \E_{\z\sim\D_{m}^{\ell}}\left(H\circ f_{\ell,\A}(\z)-H\circ F_{\ell}(\z)\right)H'(f_{\ell,\A})\frac{\partial}{\partial b_{i}}\left(\sum_{i=1}^{N}(a_{i}+b_{i})\sigma(\th_{i}^{\ell},\z)\right)\bigg|_{b_{i}=0}\\
= & \E_{\z\sim\D_{m}^{\ell}}\left(H\circ f_{\ell,\A}(\z)-H\circ F_{\ell}(\z)\right)H'(f_{\ell,\A})\sigma(\th_{i}^{\ell},\z)\\
= & r_{\A,i}.
\end{align*}
This for implementation in practice, we can introduce $b_{i}$ with
its value fixed $0$ and calculate its gradient using, which is $r_{\A,i}$
using the auto differentiate operator in common deep learning libraries.

\subsection{Details on Numeric Verification of Rate} \label{apx: rate verify}

In this section, we give details on the toy experiment on verifying
numeric rate. We first introduce the problem setup for the comparison
between pruning and direct gradient training in obtaining small network.
We use the two-hidden-layer deep mean field network formulated by
\citet{araujo2019mean}. Suppose that the second hidden layer (the one close to
output) has 50 neurons; the first hidden layer (the one close to input)
has $n\le50$ neurons with 50 dimensional feature map; and the input
has 100 dimension. That is, we consider the following deep mean field
network
\[
F^{n}(\x)=F_{2}\circ F_{1}^{n}(\x),
\]
where 
\[
F_{1}^{n}(\x)=\frac{1}{n}\sum_{i=1}^{n}a_{1,i}\text{\text{ReLU}}(\boldsymbol{b}_{1,i}^{\top}\x)
\]
with $\x\in\R^{100}$, $\boldsymbol{b}_{1,i}\in\R^{100\times50}$,
$a_{1,i}\in\R$. And 
\[
F_{2}(\z)=\frac{1}{50}\sum_{i=1}^{50}a_{2,i}\text{ReLU}(\boldsymbol{b}_{2,i}^{\top}\z),
\]
with $\z\in\R^{50}$, $\boldsymbol{b}_{2,i}\in\R^{50}$, $a_{1,i}\in\R$.
Suppose that we train the original network $F^{N}$ with $N=50$ neuron
at the first hidden layer using gradient descent defined in \citet{araujo2019mean}
for $T$ time ($T<\infty$) with random initialization. To obtain
a small network with $n$ neurons at the first hidden layer, we consider
two approaches. In the first approach, we prune the first hidden layer
of the trained $F^{N}$ using local imitation to obtain $F_{\text{local}}^{n}$
where $n$ indicates the number of neurons remained in the first hidden
layer. In the second approach, we direct train the network $F_{\text{direct\ train}}^{n}$
with $n$ neurons in the first layer using the same gradient descent
dynamics, initialization and training time as that in training $F^{N}$.
By the analysis in \citet{araujo2019mean}, we have $\DD[F_{\text{direct\ train}}^{n},F^{N}]=\mathcal{O}(n^{-1})$.
And by Theorem \ref{thm: converge_deep}, we have $\DD[F_{\text{local}}^{n},F^{N}]=\mathcal{O}(\exp(-\lambda n))$
for some $\lambda>0$. This implies that pruning is provably much
better than directly training in obtaining compact neural network.

Now we introduce the experiment settings. To simulate the data, we
first generate a random network 
\[
F_{\text{gen}}(\x)=\left(\exp\left(\boldsymbol{w}_{2}/10\right)-0.5\right)^{\top}\text{\text{Tanh}}(\text{sin}\left(2\pi\boldsymbol{w}_{1}\right)^{\top}\x/5)/1000,
\]
where $\boldsymbol{w}_{1}\in\R^{1000\times100}$ and $\boldsymbol{w}_{2}\in\R^{1000}$
is generated by randomly sampling from uniform distribution $\text{Unif}[0,1]$
(each element is sampled independently). And then we generate the
training data by sampling feature $\x$ from $\text{Unif}[0,1]$ (each
coordinate is sampled independently) and then generate label $y=F_{\text{gen}}(\x)$.
The simulated training dataset consists of 200 data points. We initialize
the parameters of $F^{N}$ and $F_{\text{direct\ train}}^{n}$ from
standard Gaussian distribution with variance 1, $\mathcal{N}(0,1)$
(each element are initialized independently) and both $F^{N}$ and
$F_{\text{direct\ train}}^{n}$ are trained using the same and sufficiently
long time to ensure convergence. We also include the pruned model
using global imitation, which is denoted as $F_{\text{global}}^{n}$.
The pruned models are not finetuned. We vary different $n$ and summarize
the discrepancy.

\subsection{Theory on Pruning All Layers} \label{apx: rate all layer}
In the main text, we mainly discuss the convergence rate of pruning
one layer. In this section, we discuss how to apply our convergence
rate for single layer pruning to obtain an overall convergence rate.
Following the layer-wise procedure introduced in Section \ref{sec:
prac}, suppose that the algorithms prunes $F_{\ell}$ to $f_{\ell,\A_{\ell}}$,
$\ell\in[L]$. And thus, during the layer-wise pruning, the algorithm
generates a sequence of pruned networks
\begin{align*}
f_{[0]} & =F_{L}\circ F_{L-1}\circ...\circ F_{3}\circ F_{2}\circ F_{1}\\
f_{[1]} & =F_{L}\circ F_{L-1}\circ...\circ F_{3}\circ F_{2}\circ f_{1,\A_{1}}\\
f_{[2]} & =F_{L}\circ F_{L-1}\circ...\circ F_{3}\circ f_{2,\A_{2}}\circ f_{1,\A_{1}}\\
 & \vdots\\
f_{[L-1]} & =F_{L}\circ f_{L-1,\A_{L-1}}\circ...\circ f_{3,\A_{3}}\circ f_{2,\A_{2}}\circ f_{1,\A_{1}}\\
f_{[L]} & =f_{L,\A_{L}}\circ f_{L-1,\A_{L-1}}\circ...\circ f_{3,\A_{3}}\circ f_{2,\A_{2}}\circ f_{1,\A_{1}}
\end{align*}
Thus here $f_{[\ell]}$ is the network with the first $\ell$ layers
pruned, $f_{[L]}$ is the final pruned network with all layers pruned
and $f_{[0]}$ is the original network. Notice that $f_{[\ell]}$
is obtained by pruning the $\ell$-th layer of $f_{[\ell-1]}$. In
this step, we suppose that we try both greedy local and global imitation
and obtain $f_{\ell,\A_{\ell}^{\text{local}}}$ and $f_{\ell,\A_{\ell}^{\text{global}}}$
with $\left\Vert \A_{\ell}^{\text{local}}\right\Vert _{0}=\left\Vert \A_{\ell}^{\text{global}}\right\Vert _{0}$.
And if $\DD[F_{L}\circ...F_{\ell+1}\circ f_{\ell,\A_{\ell}^{\text{local}}}\circ...\circ f_{1,\A_{1}},f_{[\ell-1]}]\le\DD[F_{L}\circ...F_{\ell+1}\circ f_{\ell,\A_{\ell}^{\text{global}}}\circ...\circ f_{1,\A_{1}},f_{[\ell-1]}]$,
we set $\A_{\ell}^{\text{}}=\A_{\ell}^{\text{local}}$, else we set
$\A_{\ell}^{\text{}}=\A_{\ell}^{\text{global}}$. Define $H_{[\ell]}=F_{L}\circ F_{L-1}\circ...\circ F_{\ell+1}$,
$\ell\in[L-1]$ (here $H_{[L-1]}=F_{L}$) and $\z_{[\ell]}^{(i)}=f_{\ell-1,\A_{\ell-1}}\circ...\circ f_{1,\A_{1}}(\x^{(i)})$,
$\ell\in[L-1]$ (here we define $\z_{[1]}^{(i)}=\x^{(i)}$). The set
$\D_{m}^{[\ell]}:=\left(\z_{[\ell]}^{(i)}\right)_{i=1}^{m}$ denotes
the distribution of training data pushed through the first $\ell-1$
layers.

We introduce the following assumption on the boundedness.
\begin{asm} \label{asm: bound_all}
Assume that for any $i\in[N]$, $\ell\in[L-1]$, $\z_{[\ell]}^{(j)}\in\D_{m}^{[\ell]}$,
we have $\left\Vert \sigma(\th_{i}^{\ell},\z_{[\ell]}^{(j)})\right\Vert \le c_{2}$
and $\left\Vert H_{[\ell]}\right\Vert _{\lip}\le c_{2}$ for some
$c_{2}<\infty$.
\end{asm}

\begin{thm} [Overall Convergence] \label{thm: overall}
Under assumption \ref{asm: bound_all}, we have $\sqrt{\DD[f_{[L]},F]}=\mathcal{O}\left(\sum_{\ell=1}^{L}\exp\left(-\frac{\lambda_{\ell}}{2}\left\Vert \A_{\ell}\right\Vert _{0}\right)\right)$,
with $\lambda_{\ell}>0$ for all $\ell\in[L]$ depending on $f_{[\ell-1]}$.
\end{thm}

\subsection{DGCNN Experiment} \label{apx: dgcnn}

We deploy our method on DGCNN \citep{wang2019dynamic}. DGCNN contains 4 EdgeConv layers that use K-Nearest-Neighbor(KNN) to aggregate the information from the output of convolution operation. Pruning the convolution operation in EdgeConv can significantly speed up the KNN operation and therefor make the whole model more computational efficient.

\paragraph{Settings} The full network is trained with SGD optimizer with momentum 0.9 and weight decay $1 \times 10^{-4}$. We train the model using 64 batch size with an initial learning rate 0.1 for 250 epochs. We apply cosine learning rate scheduler during the training and decrease the learning rate to 0.001 at the final epoch. During the pruning, we use 32 batch sizes and the other settings keep the same as our ImageNet experiment in Section \ref{sec: imagenet}.

\section*{Technical Lemmas}
We introduce several technical Lemmas that are useful for proving the main theorems.

\begin{lem} \label{techlem: rl_equ}
Given some convex set $M\subset\R^{d}$, for any $\boldsymbol{q}_{1}\in\text{ri}M$
and $\boldsymbol{q}_{2}\in\text{cl}M$. Then all the points from the
half-segment $[\boldsymbol{q}_{1},\boldsymbol{q}_{2})$ belongs to
the relative interior of $M$, i.e.,
\[
[\boldsymbol{q}_{1},\boldsymbol{q}_{2})\hat{=}\left\{ (1-\lambda)\boldsymbol{q}_{1}+\lambda\boldsymbol{q}_{2}\mid0\le\lambda<1\right\} \subseteq\text{ri}M.
\]
\end{lem}

\begin{lem} \label{techlem: rl_nonempty}
Let $M$ be a convex set in $\R^{d}$, then if $M$ is nonempty, then
the relative interior of $M$ is nonempty.
\end{lem}

Lemma \ref{techlem: rl_equ} and \ref{techlem: rl_nonempty} are classic results from convex optimization.

\begin{lem} \label{techlem: rl_avg}
Define 
\[
M=\text{conv}\left\{ \boldsymbol{q}\mid\boldsymbol{q}\in S\right\} ,
\]
where $S=\{\boldsymbol{q}_{1},...,\boldsymbol{q}_{n}\}\subseteq\R^{d}$
with $1\le n<\infty$. Define $\bar{\boldsymbol{q}}=\frac{1}{n}\sum_{i=1}^{n}\boldsymbol{q}_{i}$,
then $\bar{\boldsymbol{q}}\in\text{ri}M$.
\end{lem}

\begin{lem} \label{techlem: rl_ine}
Suppose that for some $\lambda>0$ such that $\left(\mathcal{B}(\bar{\h},\lambda)\cap\text{Aff}\M\right)\subseteq\M$,
then $\max_{\boldsymbol{s}\in\M}\left\langle \bar{\h}-\h_{A},\bar{\h}-\boldsymbol{s}\right\rangle \ge\lambda\left\Vert \bar{\h}-\h_{A}\right\Vert $.
\end{lem}

\begin{lem} \label{techlem: diameter}
Under Assumption \ref{asm: bound}, for any $\h,\h'\in\M$, $\left\Vert \h-\h'\right\Vert \le D$
for some $D\le2\sqrt{m}c_{1}$. Here $D$ can be viewed as the diameter of $\M$.
\end{lem}

\begin{lem} \label{techlem: FW_des}
Under assumption \ref{asm: bound}, suppose $\tilde{\boldsymbol{s}}_{k}^{*}=\underset{\boldsymbol{s}\in\M}{\arg\min}\left\langle \bar{\h}-\h_{\A(k)},\boldsymbol{s}-\h_{\A(k)}\right\rangle$  and $\tilde{\gamma}_{k}^{*}=\underset{\gamma\in[0,1]}{\arg\min}\left\Vert \h_{\A(k)}+\gamma\left(\tilde{\boldsymbol{s}}_{k}^{*}-\h_{\A(k)}\right)-\bar{\h}\right\Vert ^{2}$, then 
\[
\left\Vert \h_{\A(k)}+\tilde{\gamma}_{k}^{*}\left(\tilde{\boldsymbol{s}}_{k}^{*}-\h_{\A(k)}\right)-\bar{\boldsymbol{h}}\right\Vert ^{2}\le\rho\left\Vert h_{A(k)}-\bar{h}\right\Vert ^{2},
\]
for some $\rho\in(0,1)$.

\end{lem}

\begin{lem} \label{techlem: num_seq}
Consider the following number sequence $x_{k+1}^{2}\le ax_{k}^{2}-bx_{k}+c$.
Suppose that this number sequence satisfies the following conditions:
(1) $a>1$, $b\ge0$, $c\ge0$; (2) $x_{k}\ge0$ for any $k$; (3)
$(a-1)x^{2}-bx^{2}+c$ has two real roots $z_{1}\le z_{2}$; (we allow
$z_{1}=z_{2}$); (4) $\sqrt{c}\le z_{2}$; (5) $x_{0}\le z_{2}$.
Then $\sup_{k}x_{k}\le z_{2}$.
\end{lem}

\section*{Proof of Main Theorems}

\subsubsection{Proof of Theorem \ref{thm: converge_deep}}

Using Lemma \ref{techlem: rl_avg}, we know that $\bar{\h}\in\text{ri}\M$,
which indicate that there exists some $\lambda>0$ such that 
\[
\mathcal{B}(\bar{\h},\lambda) \cap \text{Aff} \M \subseteq\M,
\]
where $\mathcal{B}(\bar{\h},\lambda)$ denotes the ball with radius
$\lambda$ centered at $\bar{\h}$. Define $\text{Extre}(\M)$ as
the set of extreme points of $\M$, we know that $\text{Extre}(\M)\subseteq\{\h_{1},...,\h_{N}\}$.
Consider the following problem 
\[
\min_{\boldsymbol{s}\in\M}\left\langle \bar{\h}-\h_{\A(k)},\boldsymbol{s}-\h_{\A(k)}\right\rangle .
\]
As the objective $\left\langle \bar{\h}-\h_{\A(k)},\boldsymbol{s}-\h_{\A(k)}\right\rangle $
is linear w.r.t. $\boldsymbol{s}$, we know that $\boldsymbol{s}\in\text{Extre}(\M)\subseteq\{\h_{1},...,\h_{N}\}$.
Also, for any $i\in[N]$, we have $[0,1]\subseteq U_{i}$. This gives
that
\begin{align*}
\min_{i\in[N]}\min_{\gamma\in U_{i}}\bar{\DD}[(1-\gamma)f_{\ell,\A(k)}+\gamma\sigma(\th_{i},\cdot),F_{\ell}] & \le\min_{\gamma\in[0,1]}\left\Vert \h_{\A(k)}+\gamma\left(\tilde{\boldsymbol{s}}_{k}^{*}-\h_{\A(k)}\right)-\bar{\h}\right\Vert ^{2}\\
 & \le(1-\lambda^{2}/D^{2})\left\Vert \h_{\A(k)}-\bar{\h}\right\Vert ^{2},
\end{align*}
where $\tilde{\boldsymbol{s}}_{k}^{*}=\arg\min_{\boldsymbol{s}\in\M}\left\langle \bar{\h}-\h_{\A(k)},\boldsymbol{s}-\h_{\A(k)}\right\rangle $.
Here the last inequality is by Lemma \ref{techlem: FW_des}. This
gives that 
\[
\left\Vert \h_{\A(k+1)}-\bar{\h}\right\Vert ^{2}\le(1-\lambda^{2}/D^{2})\left\Vert \h_{\A(k)}-\bar{\h}\right\Vert ^{2}.
\]
And thus we have 
\[
\left\Vert \h_{\A(k)}-\bar{\h}\right\Vert ^{2}\le(1-\lambda^{2}/D^{2})^{k}\left\Vert \h_{\A(0)}-\bar{\h}\right\Vert ^{2}.
\]
And thus we have 
\[
\DD[f_{\A(k)},F]\le c_{1}^{2}(1-\lambda^{2}/D^{2})^{k}\left\Vert \h_{\A(0)}-\bar{\h}\right\Vert ^{2}\le c_{1}^{2}(1-\lambda^{2}/D^{2})^{\left\Vert \A(k)\right\Vert _{0}}\left\Vert \h_{\A(0)}-\bar{\h}\right\Vert ^{2},
\]
where the last inequality is by $\left\Vert \A(k)\right\Vert _{0}\le k$.

\subsection*{Proof of Theorem \ref{thm: solve_greedy_fast}}
Notice that if we have $\frac{\left\langle \bar{\h}-\h_{\A(k)},\h_{i_{k}^{*}}-\h_{\A(k)}\right\rangle }{\left\Vert \h_{i_{k}^{*}}-\h_{\A(k)}\right\Vert ^{2}}\ge1$,
then $\gamma_{k}^{*}=1$ and in this case, 
\[
\left\Vert \h_{\A(k+1)}-\bar{\h}\right\Vert =\bar{\DD}[\sigma(\th_{i_{k}^{*}},\cdot),F(\cdot)]\ge\bar{\DD}[\sigma(\th_{i_{0}^{*}},\cdot),F(\cdot)].
\]
On the other hand, since $0<\left\Vert \bar{\h}-\h_{\A(k)}\right\Vert $,
by the argument in proving Theorem \ref{thm: converge_deep}, we have 
\[
\left\Vert \bar{\h}-\h_{\A(k+1)}\right\Vert \le\sqrt{1-\lambda^{2}/D^{2}}\left\Vert \bar{\h}-\h_{\A(k)}\right\Vert <\left\Vert \bar{\h}-\h_{\A(k)}\right\Vert .
\]
This gives that 
\[
\left\Vert \bar{\h}-\h_{\A(k+1)}\right\Vert <\left\Vert \bar{\h}-\h_{\A(k)}\right\Vert \le\bar{\DD}[\sigma(\th_{i_{0}^{*}},\cdot),F(\cdot)],
\]
which makes contradiction.

\subsection*{Proof of Theorem \ref{thm: local_fixstep_converge}}
Using Lemma \ref{techlem: rl_avg}, we know that $\bar{\h}\in\text{ri}\M$,
which indicate that there exists some $\lambda>0$ such that 
\[
\mathcal{B}(\bar{\h},\lambda) \cap \text{Aff} \M \subseteq\M,
\]
where $\mathcal{B}(\bar{\h},\lambda)$ denotes the ball with radius
$\lambda$ centered at $\bar{\h}$. Following the same argument of \citet{ye2020good} in proving theorem 2, we have $\left\Vert \bar{\h}-\h_{\A(k)}\right\Vert ^{2}=\mathcal{O}((k+1)^{-2})$. The result that $\left\Vert \A(k)\right\Vert _{0}\le k+1$ is obvious as in each iteration, the number of nonzero elements in $\A$ at most increases by 1.

\subsection*{Proof of Theorem \ref{thm: global_converge}}
Suppose that at iteration $k$, we have $h_{\A(k)}$. And the global
imitation algorithm returns $h_{\A(k+1)}$ with $f_{\ell.\A(k+1)}=H_{2}\circ\left[(1-\gamma_{k})f_{\ell,\A(k)}+\sigma(\th_{i_{k}^{*}},\cdot)\right]\circ H_{1}$.
We also define $i'_{k}$ as the solution of local
imitation. And we let $f_{\A'(k+1)}=H_{2}\circ\left[(1-\gamma_{k})f_{\ell,\A(k)}+\sigma(\th_{i'_{k}},\cdot)\right]\circ H_{1}$.
Define $\w_{k+1}=(k+1)(\bar{\h}-\h_{\A(k)})$, $\w'_{k+1}=(k+1)\left(\h-\h_{\A'(k)}\right)$, $\W_{k+1}=(k+1)\left(H\circ\bar{\h}-H\circ \h_{\A(k)}\right)$ and $\W'_{k+1}=(k+1)\left(H\circ\bar{\h}-H\circ \h_{\A'(k)}\right)$.
We have 
\begin{align*}
\left\Vert \W_{k+1}\right\Vert ^{2} & \le\left\Vert \W'_{k+1}\right\Vert ^{2}\\
 & =(k+1)^{2}\left\Vert H\circ\bar{\h}-H\circ \h_{\A'(k)}\right\Vert ^{2}\\
 & \overset{(1)}{\le}\kappa_{1}^{2}(k+1)^{2}\left\Vert \bar{\h}-\h_{\A'(k)}\right\Vert ^{2}\\
 & =\kappa_{1}^{2}\left\Vert \w'_{k+1}\right\Vert ^{2}\\
 & \overset{(2)}{\le}\kappa_{1}^{2}\left(\left\Vert \w{}_{k}\right\Vert ^{2}-2\lambda\left\Vert \w{}_{k}\right\Vert +D^{2}\right)\\
 & \overset{(3)}{\le}\kappa_{1}^{2}\left(\kappa_{2}^{2}\left\Vert \W{}_{k}\right\Vert ^{2}-\kappa_{2}2\lambda\left\Vert \W{}_{k}\right\Vert +D^{2}\right)\\
 & =\kappa_{1}^{2}\kappa_{2}^{2}\left\Vert \W{}_{k}\right\Vert ^{2}-2\kappa_{1}^{2}\kappa_{2}\lambda\left\Vert \W{}_{k}\right\Vert +\kappa_{1}^{2}D^{2}.
\end{align*}
Here $D$ is the quantities defined in Lemma \ref{techlem: diameter}, $(1)$ and $(3)$ use the definition of $\kappa_{1}$ and $\kappa_{2}$
and (2) is by the argument of \citet{ye2020good} in proving Theorem 2 (notice that their argument also applies to the case that $\bar{\h}$ is in the relative interior of $\M$, which is proved by Lemma \ref{techlem: rl_avg}, instead of that $\bar{\h}$ is in the interior of $\M$). By the
assumption that $D^{2}\ge\kappa_{1}^{2}\kappa_{2}^{2}(D^{2}-\lambda^{2}),$
the formula $\kappa_{1}^{2}\kappa_{2}^{2}x^{2}-2\kappa_{1}^{2}\kappa_{2}\lambda x+\kappa_{1}^{2}D^{2}=x^{2}$
has two real root, denoted by $z_{1}\le z_{2}$, where 
\begin{align*}
z_{1} & =\frac{\kappa_{1}^{2}\kappa_{2}\lambda-\kappa_{1}\sqrt{\kappa_{1}^{2}\kappa_{2}^{2}(\lambda^{2}-D^{2})+D^{2}}}{(\kappa_{1}^{2}\kappa_{2}^{2}-1)}\\
z_{2} & =\frac{\kappa_{1}^{2}\kappa_{2}\lambda+\kappa_{1}\sqrt{\kappa_{1}^{2}\kappa_{2}^{2}(\lambda^{2}-D^{2})+D^{2}}}{(\kappa_{1}^{2}\kappa_{2}^{2}-1)}.
\end{align*}
We define $q_{1}=\kappa_{1}^{2}\kappa_{2}^{2}$ and $q_{2}=\kappa_{1}^{2}\kappa_{2}$,
and we have 
\[
\left\Vert \W_{k+1}\right\Vert ^{2}\le q_{1}\left\Vert \W{}_{k}\right\Vert ^{2}-2q_{2}\lambda\left\Vert \W{}_{k}\right\Vert +\kappa_{1}^{2}D^{2}.
\]
If $q_{1}=1$, then the rate holds by directly applying the argument
of \citet{ye2020good} in proving Theorem 2. If $q_{1}>1$, we know that $2q_{2}\lambda\ge0$
and $\kappa_{1}^{2}D^{2}\ge0$; $\left\Vert \W{}_{k}\right\Vert \ge0$
for any $k$ by its definition; the formula $q_{1}x^{2}-2q_{2}\lambda x+\kappa_{1}^{2}D^{2}=x^{2}$
has two real roots $z_{1}\le z_{2}$; $z_{2}\ge\kappa_{1}D$ by the
assumption; and $\left\Vert \W_{k+1}\right\Vert \le z_{2}$ by the
assumption. Using Lemma \ref{techlem: num_seq}, we have, for any $k$,
\[
\left\Vert \W_{k}\right\Vert \le z_{2},
\]
which implies that 
\[
\left\Vert H\circ\bar{\h}-H\circ \h_{\A(k)}\right\Vert ^{2}=\mathcal{O}((k+1)^{-2}),
\]
and thus $\DD[f_{\A(k)},F]=\mathcal{O}(k^{-2})$. The result that
$\left\Vert \A(k)\right\Vert _{0}\le k+1$ is obvious as in each iteration,
$\left\Vert \A(k)\right\Vert _{0}$ at most increase 1.

\subsection*{Proof of Theorem \ref{thm: overall}}
When pruning the $\ell$-th layer, if this layer is pruned
by local imitation, by applying Theorem \ref{thm: converge_deep} on the $\ell$-th layer of $f_{[\ell-1]}$, we have 
\[
\sqrt{\DD[f_{[\ell]},f_{[\ell-1]}]}=\mathcal{O}\left(\exp(-\frac{\lambda_{\ell}}{2}\left\Vert \A_{\ell}\right\Vert _{0})\right),
\]
for some $\lambda_{\ell}>0$. Else if this layer is pruned by global
imitation, we have 
\begin{align*}
\sqrt{\DD[f_{[\ell]},f_{[\ell-1]}]} & \le\sqrt{\DD[F_{L}\circ...F_{\ell+1}\circ f_{\ell,\A_{\ell}^{\text{local}}}\circ...\circ f_{1,\A_{1}},f_{[\ell-1]}]}\\
 & =\mathcal{O}\left(\exp(-\frac{\lambda_{\ell}}{2}\left\Vert \A_{\ell}^{\text{local}}\right\Vert _{0})\right)=\mathcal{O}\left(\exp(-\frac{\lambda_{\ell}}{2}\left\Vert \A_{\ell}^{\text{}}\right\Vert _{0})\right).
\end{align*}
Using triangle inequality, we know that 
\[
\sqrt{\DD[f_{[L]},F]}\le\sum_{\ell=1}^{L}\sqrt{\DD[f_{[\ell]},f_{[\ell-1]}]}=\mathcal{O}\left(\sum_{\ell=1}^{L}\exp\left(-\frac{\lambda_{\ell}}{2}\left\Vert \A_{\ell}\right\Vert _{0}\right)\right),
\]
with $\lambda_{\ell}>0$ for all $\ell\in[L]$.

\section*{Proof of Technical Lemmas}

\subsection*{Proof of Lemma \ref{techlem: rl_avg}}
The case that $n=1$ is trivial and we consider the case that $n\ge2$.
By the definition, we know that $M$ is an non-empty and closed convex
set. And thus by Lemma \ref{techlem: rl_nonempty}, $\text{ri}M$
is not empty. Define 
\[
\tilde{\boldsymbol{q}}\in\text{ri}M,\ \tilde{\boldsymbol{q}}=\sum_{i=1}^{n}\alpha_{i}\boldsymbol{q}_{i},\ \ \ \sum_{i=1}^{n}\alpha_{i}=1\ \text{and}\ \alpha_{i}\ge0\ \forall i\in[n].
\]
We define $\alpha_{\max}=\max_{i\in[n]}\alpha_{i}$. Notice that $\alpha_{\max}\ge1/n$,
otherwise, if $\alpha_{\max}<1/n$, we have $\sum_{i=1}^{n}\alpha_{i}\le n\alpha_{\max}<1$,
which makes contradiction. If $\alpha_{\max}=1/n$, then $\alpha_{i}=1/n$
for all $i\in[n]$, otherwise, $\sum_{i=1}^{n}\alpha_{i}<1$, which
makes contradiction. In the case that $\alpha_{\max}=1/n$, we have
already obtained the desired result.

Now we assume $\alpha_{\max}>\frac{1}{n}$. Define $\lambda=1-\frac{1}{n\alpha_{\max}}\in[0,1)$
and $\beta_{i}=\frac{\alpha_{\max}-\alpha_{i}}{n\alpha_{\max}-1}.$
Notice this gives that 
\[
\sum_{i=1}^{n}\beta_{i}=\sum_{i=1}^{n}\frac{\alpha_{\max}-\alpha_{i}}{n\alpha_{\max}-1}=\frac{n\alpha_{\max}-\sum_{i=1}^{n}\alpha_{i}}{n\alpha_{\max}-1}=1\ \ \ \text{and}\ \beta_{i}\ge0\ \forall i\in[n].
\]
We define $\boldsymbol{q}'=\sum_{i=1}^{n}\beta_{i}\boldsymbol{q}_{i}$
and by the property of $\beta_{i}$ and the definition of $M$, we
have $\boldsymbol{q}'\in M=\text{cl}M$. Notice that 
\[
\bar{\boldsymbol{q}}=\frac{1}{n}\sum_{i=1}^{n}\boldsymbol{q}_{i}=(1-\lambda)\sum_{i=1}^{n}\alpha_{i}\boldsymbol{q}_{i}+\lambda\sum_{i=1}^{n}\beta_{i}.
\]
Using Lemma \ref{techlem: rl_equ}, we know that $\bar{\boldsymbol{q}}\in\text{ri}M$.

\subsection{Proof of Lemma \ref{techlem: rl_ine}}

Notice that by choosing $s'=\bar{\h}-\lambda\frac{\bar{\h}-\h_{A}}{\left\Vert \bar{\h}-\h_{A}\right\Vert }\in\M$,
we have 
\[
\max_{s\in\M}\left\langle \bar{\h}-\h_{A},\bar{\h}-\boldsymbol{s}\right\rangle \ge\left\langle \bar{\h}-\h_{A},\bar{\h}-\boldsymbol{s}'\right\rangle =\lambda\left\Vert \bar{\h}-\h_{A}\right\Vert .
\]

\subsection{Proof of Lemma \ref{techlem: diameter}}

Notice that for any $i\in[N]$, 
\[
\left\Vert \h_{i}\right\Vert =\sqrt{\sum_{j=1}^{m}\sigma^{2}(\th_{i},\z^{(j)})}\le\sqrt{m}c_{1}.
\]
And for any $\h\in\M$, we have $\h=\sum_{i=1}^{N}\beta_{i}\h_{i}$,
for some $\beta_{i}\ge0$ and $\sum_{i=1}^{N}\beta_{i}=1$, which
gives that 
\[
\left\Vert \h\right\Vert =\left\Vert \sum_{i=1}^{N}\beta_{i}\h_{i}\right\Vert \le\sum_{i=1}^{n}\beta_{i}\left\Vert \h_{i}\right\Vert \le\sqrt{m}c_{1}.
\]

\subsection*{Proof of Lemma \ref{techlem: FW_des}}

Proof of this Lemma follows standard argument in analyzing Frank Wolfe
algorithm. We include it for the completeness. Notice that
\[
\tilde{\boldsymbol{s}}_{k}^{*}=\underset{\boldsymbol{s}\in\M}{\arg\min}\left\langle \bar{\h}-\h_{\A(k)},\boldsymbol{s}-\h_{\A(k)}\right\rangle =\underset{\boldsymbol{s}\in\M}{\arg\min}\left\langle \bar{\h}-\h_{\A(k)},\boldsymbol{s}-\bar{\h}\right\rangle =-\underset{\boldsymbol{s}\in\M}{\arg\max}\left\langle \bar{\h}-\h_{\A(k)},\bar{\h}-\boldsymbol{s}\right\rangle .
\]
Using Lemma \ref{techlem: rl_ine}, we know that $\left\langle \bar{\h}-\h_{\A(k)},\bar{\h}-\tilde{\boldsymbol{s}}_{k}^{*}\right\rangle \le-\lambda\left\Vert \bar{\h}-\h_{\A(k)}\right\Vert $.
Notice that 
\begin{align*}
 & \left\langle \bar{\h}-\h_{\A(k)},\tilde{\boldsymbol{s}}_{k}^{*}-\h_{\A(k)}\right\rangle \\
= & \left\langle \bar{\h}-\h_{\boldsymbol{A}(k)},\bar{\h}-\h_{\A(k)}+\tilde{\boldsymbol{s}}_{k}^{*}-\bar{\h}\right\rangle \\
= & -\left\langle \bar{\h}-\h_{\A(k)},\bar{\h}-\tilde{\boldsymbol{s}}_{k}^{*}\right\rangle +\left\Vert \bar{\h}-\h_{\A(k)}\right\Vert ^{2}\\
\le & -2\left\langle \bar{\h}-\h_{\A(k)},\bar{\h}-\tilde{\boldsymbol{s}}_{k}^{*}\right\rangle +\left\Vert \bar{\h}-\h_{\A(k)}\right\Vert ^{2}+\left\Vert \bar{\h}-\tilde{\boldsymbol{s}}_{k}^{*}\right\Vert ^{2}\\
= & \left\Vert \left(\bar{\h}-\h_{\A(k)}\right)-\left(\bar{\h}-\tilde{\boldsymbol{s}}_{k}^{*}\right)\right\Vert ^{2}=\left\Vert \h_{\A(k)}-\tilde{\boldsymbol{s}}_{k}^{*}\right\Vert ^{2},
\end{align*}
where the last inequality uses the fact that $\left\langle \bar{\h}-\h_{\A(k)},\bar{\h}-\tilde{\boldsymbol{s}}_{k}^{*}\right\rangle \le0$.
This gives that $0\le\left\langle \bar{\h}-\h_{\A(k)},\tilde{\boldsymbol{s}}_{k}^{*}-\h_{\A(k)}\right\rangle \le\left\Vert \h_{\A(k)}-\tilde{\boldsymbol{s}}_{k}^{*}\right\Vert^2$.
And thus we have 
\begin{align*}
 & \min_{\gamma\in[0,1]}\left\Vert \h_{\A(k)}-\bar{\h}\right\Vert ^{2}-2\gamma\left\langle \bar{\h}-\h_{\A(k)},\tilde{\boldsymbol{s}}_{k}^{*}-\h_{\A(k)}\right\rangle +\gamma^{2}\left\Vert \tilde{\boldsymbol{s}}_{k}^{*}-\h_{\A(k)}\right\Vert ^{2}\\
= & \left\Vert \h_{\A(k)}-\bar{\h}\right\Vert ^{2}-\frac{\left\langle \bar{\h}-\h_{\A(k)},\tilde{\boldsymbol{s}}_{k}^{*}-\h_{\A(k)}\right\rangle ^{2}}{\left\Vert \h_{\A(k)}-\tilde{\boldsymbol{s}}_{k}^{*}\right\Vert ^{2}}\\
\le & \left\Vert \h_{\A(k)}-\bar{\h}\right\Vert ^{2}-\lambda^{2}\frac{\left\Vert \bar{\h}-\h_{\A(k)}\right\Vert ^{2}}{\left\Vert \h_{\A(k)}-\tilde{\boldsymbol{s}}_{k}^{*}\right\Vert ^{2}}\\
\le & (1-\lambda^{2}/D^{2})\left\Vert \h_{\A(k)}-\bar{\h}\right\Vert ^{2},
\end{align*}
where the last inequality is by Lemma \ref{techlem: diameter}.

\subsection*{Proof of Lemma \ref{techlem: num_seq}}
Define $f(x)=ax^{2}-bx+c$. By assumption (1) and assumption (3),
for any $z\in[z_{1},z_{2}]$, $f(z)-z^{2}\le0$. We proof the desired
result by induction. Suppose that $x_{k}\in[0,z_{2}]$. Case 1: $x_{k}\in[z_{1},z_{2}]$
and in this case, 
\[
x_{k+1}^{2}\le f(x_{k})\le x_{k}^{2}\le z_{2}^{2}.
\]
Case 2: $x_{k}\in[0,z_{1})$ and in this case 
\[
x_{k+1}^{2}\le f(x_{1})\le\max_{z\in[0,z_{1}]}f(z)\le\max(f(0),f(z_{1}))=\max(c,z_{1}^{2}).
\]
This gives that $x_{k+1}\le\max(\sqrt{c},z_{1})\le\max(z_{2},z_{1})=z_{2}.$
The desired result follows by induction.

\end{document}